\documentclass[pdflatex,sn-mathphys-num]{sn-jnl} 
\usepackage{graphicx}%
\usepackage{multirow}%
\usepackage{amsmath,amssymb,amsfonts}%
\usepackage{amsthm}%
\usepackage{mathrsfs}%
\usepackage[title]{appendix}%
\usepackage{xcolor}%
\usepackage{textcomp}%
\usepackage{manyfoot}%
\usepackage{booktabs}%
\usepackage{algorithm}%
\usepackage{algorithmicx}%
\usepackage{algpseudocode}%
\usepackage{listings}%

\usepackage{subcaption}
\usepackage{pgfplots}
\pgfplotsset{compat=newest}
\theoremstyle{thmstyleone}%
\theoremstyle{thmstyletwo}%
\theoremstyle{thmstylethree}%

\begin{document}

\title[Article Title]{Natural Spectral Fusion: \emph{p}-Exponent Cyclic Scheduling and Early Decision-Boundary Alignment in First-Order Optimization}


\author[1]{\fnm{Zhang} \sur{Gongyue}}\email{gongyuezhang@outlook.com}

\author*[2]{\fnm{Liu} \sur{Honghai}}\email{honghai.liu@icloud.com}

\abstract{
Spectral behaviors have been widely discussed in machine learning, yet the optimizer’s own spectral bias remains unclear. We argue that first-order optimizers exhibit an intrinsic frequency preference that significantly reshapes the optimization path. To address this, we propose \textbf{Natural Spectral Fusion (NSF)}: reframing training as \emph{controllable spectral coverage and information fusion} rather than merely scaling step sizes. NSF has two core principles: (1) treating the optimizer as a \emph{spectral controller} that dynamically balances low- and high-frequency information; and (2) periodically reweighting frequency bands at negligible cost, without modifying the model, data, or training pipeline. We realize NSF via a \textbf{p-exponent extension} of the second-moment term, enabling both positive and negative exponents, and implement it through \textbf{cyclic scheduling}. Theory and experiments show that adaptive methods emphasize low frequencies, SGD is near-neutral, and negative exponents amplify high-frequency information. Cyclic scheduling broadens spectral coverage, improves cross-band fusion, and induces \emph{early decision-boundary alignment}---where accuracy improves even while loss remains high. Across multiple benchmarks, with identical learning-rate strategies and fixed hyperparameters, p-exponent cyclic scheduling consistently reduces test error and test loss; on some tasks, it matches baseline accuracy with only one-quarter of the training cost. Overall, NSF reveals the optimizer’s role as an active spectral controller and provides a unified, controllable, and efficient framework for first-order optimization.}




\maketitle

\section{Introduction}\label{sec1}

Recent studies on spectral behavior (e.g., F-Principle / spectral bias) repeatedly show that neural networks tend to fit low-frequency components first\cite{xu2019frequency}\cite{rahaman2019spectral}. Some works further interpret optimization algorithms (e.g., momentum) in the spectral domain or apply spectral operations to gradients and features, indicating that training is not entirely spectrum-neutral\cite{su2016differential}. However, the mainstream view still regards the optimizer as a \emph{passive} executor: its \emph{intrinsic} spectral preference is rarely characterized, and there is no unified framework for \emph{optimizer spectral control}\cite{wilson2017marginal}\cite{zhang2024transformers}. From an optimizer-centric perspective, our key observation is that different first-order optimization methods exhibit \textbf{specific spectral preferences}, which significantly reshape optimization trajectories\cite{keskar2017improving}\cite{gupta2021adam}.

Prior work falls into three main strands. (i) \emph{Spectral bias / F-Principle:} extensive results show a low-to-high frequency fitting order, explained through Fourier, NTK, and linearization views, and linked to model depth/width, regularization, and data distributions\cite{basri2020frequency}\cite{chen2020generalized}. (ii) \emph{Optimizer-related spectral interpretations:} momentum and adaptive methods are often viewed as time--frequency filters, suppressing high-frequency noise and emphasizing smooth components; some methods also directly apply spectral filtering or enhancement to gradients and features\cite{zhang2019yellowfin}\cite{cao2021towards}. (iii) \emph{Scheduling methods:} cosine, cyclical, and related schedules improve convergence and generalization by adjusting training cadence, but remain amplitude-centric and \emph{do not explicitly control the relative weights of different spectral bands}\cite{loshchilov2017sgdr}\cite{smith2017cyclical}\cite{ smith2019super}. Overall, the field still lacks an \emph{optimizer-centric} framework that characterizes spectral tendencies and leverages them for cross-band fusion.

We propose \textbf{Natural Spectral Fusion (NSF)}: an optimizer-centric principle that organizes training as \emph{controllable spectral coverage and fusion}, rather than as a one-dimensional sequence of larger/smaller step sizes. The intuition is to modulate the optimizer’s “attention strength” across spectral bands so that training trajectories can \emph{selectively amplify and naturally fuse} low- and high-frequency components, thereby improving generalization efficiency while maintaining stability. Compared with existing learning-rate or momentum schedules, NSF no longer focuses solely on update magnitudes, but directly regulates the relative weights of different spectral bands, thus offering a new dimension of optimization. It can serve both as an independent lens to understand optimizer behavior and as a unifying framework to design spectrum-oriented scheduling strategies. In addition, NSF reveals an extra phenomenon, namely \emph{early decision-boundary alignment}---where accuracy improves even while loss remains relatively high.

Building on NSF, we propose \textbf{\(p\)-exponent cyclic scheduling}. The core idea is to extend the exponent on the second-moment term to both \emph{positive and negative} values of \(p\), and to periodically switch between them during training. In this way, the optimizer alternately emphasizes low- and high-frequency components, achieving a \emph{broader spectral coverage and natural cross-band fusion}. Unlike traditional strategies that primarily adjust step sizes or momentum, \(p\)-exponent scheduling acts directly in the spectral domain, enabling training to move beyond a single frequency preference and toward a more comprehensive spectral perspective. This mechanism is independent of model architecture and data characteristics, can serve as a general extension of first-order optimizers, and is easily integrated into existing training pipelines while offering a new perspective for spectrum-oriented optimization design.

From this perspective, the central contributions of this paper are:
\begin{itemize}
	\item \textbf{Concept (NSF).} We propose \emph{Natural Spectral Fusion (NSF)}, enabling the optimizer itself to realize \emph{controllable spectral coverage and fusion}, and providing a new optimization dimension beyond architecture, data, and loss design.
	\item \textbf{Method (\(p\)-Exponent Cyclic Scheduling).} We extend the second-moment exponent to both positive and negative values of \(p\), and introduce a cyclic scheduling strategy that alternately emphasizes different spectral bands, thereby achieving \emph{broader spectral coverage and natural cross-band fusion}.
	\item \textbf{Theory \& Evidence.} We derive spectral dynamics and characterize the interactions across frequency bands; we further predict and verify that \(p>0\) favors low frequencies (low-pass), \(p<0\) favors high frequencies (high-pass), and \(p=0\) is near-neutral. In both theoretical demonstrations and real tasks, we show broadened, optimizer-specific spectral coverage and the phenomenon of early decision-boundary alignment, which lead to improved efficiency and reduced test error/loss.
\end{itemize}

\section{Related Work and Background}

This section surveys spectral phenomena in deep learning, spectral interpretations of optimizers, frequency-oriented strategies on the data/feature side, and mainstream scheduling methods.

\subsection{Spectral Phenomena in Deep Learning}

A large body of research has shown that deep neural networks typically exhibit a ``low-to-high'' fitting order during training. Specifically, networks tend to capture smooth, low-oscillation components---that is, low-frequency information---at early stages, and only gradually cover more complex, fine-grained high-frequency details as training progresses. This regularity is widely referred to as the \emph{spectral bias} or the \emph{F-Principle}.

This phenomenon has been validated across a variety of modalities (e.g., images, speech, and text), network scales (from shallow to very deep models), and task settings. Explanations have been proposed from multiple theoretical perspectives. Some studies employ Fourier decompositions to reveal differences in learning speed across frequencies. Others build on the neural tangent kernel (NTK) and linearized dynamics to connect parameter evolution with frequency coverage. Further interpretations invoke implicit regularization and loss-landscape properties to account for the network’s natural preference toward low-frequency components.

The order of frequency fitting is influenced by many factors. Architectural choices such as network depth and width, activation functions, and normalization schemes can affect the relative pace of learning different components. Data distribution and regularization methods (e.g., weight decay, dropout) also shape spectral behavior. Moreover, noise levels and preprocessing procedures contribute to the dynamics. In certain nonstationary settings or under strong data augmentation, the canonical ``low-to-high'' order may deviate or even interleave across frequencies.

\subsection{Optimizers as Spectral Filters}

Another line of research seeks to interpret optimization algorithms through a spectral perspective. These works generally argue that different optimizers exert selective influences on frequency components during training, and that such influences are not spectrum-neutral. A typical conclusion is that momentum-based methods act as time–frequency smoothers and denoisers: they effectively suppress high-frequency noise in the gradients, thereby stabilizing the optimization trajectory, while simultaneously accumulating and preserving low-frequency components more effectively. In contrast, plain stochastic gradient descent often exhibits a more direct and unfiltered spectral response.

Adaptive second-moment methods (such as the Adam family) further modify the relative weighting of different frequency bands. By accumulating and rescaling squared gradients, these methods implicitly adjust the pace of updates across frequencies. This mechanism alters the optimizer’s behavior at different stages of training, manifesting as amplification or suppression of certain spectral components. As a result, adaptive methods implicitly regulate the relative importance of frequencies, which in turn shapes both convergence paths and generalization properties.

In addition, some approaches directly manipulate gradients or intermediate features in the frequency domain. For example, several studies propose applying low-pass or band-pass filtering to gradients in order to eliminate high-frequency perturbations; others employ spectral enhancement or denoising techniques to highlight critical frequency information. Such methods are often motivated by robustness and generalization, and tend to be especially effective in noisy environments or tasks with prominent frequency-domain characteristics.

\subsection{Spectral Control via Data and Features (Non-Optimizer Side)}

In deep learning practice, many efforts have sought to guide spectral behavior from the \emph{data or feature} side. Such methods include frequency-domain augmentation and regularization, Fourier feature mappings, band-limited training, and explicit suppression or enhancement of particular bands. For example, spectral augmentation can artificially amplify or attenuate certain frequency components in training data to steer the model’s attention, while Fourier feature mappings inject periodic components into the input space to improve the learning of high-frequency signals.

Domain-specific applications often provide further motivation for spectral strategies. In medical imaging, low-frequency information usually corresponds to global tissue structure, whereas high-frequency components carry critical details such as lesion boundaries. In speech recognition, most energy is concentrated in mid-to-low frequency ranges, while high frequencies influence timbre and clarity. In remote sensing, different frequency components capture textures, boundaries, and periodic patterns of land surfaces. Accordingly, many application-driven methods incorporate prior spectral preferences into training data or representations to improve task performance.

Nevertheless, these methods predominantly rely on \emph{data selection, feature engineering, or architectural design}. In other words, they shape spectral distributions indirectly through external conditions, but they do not directly control how the optimizer allocates its attention across spectral bands. This limitation leaves open the possibility that an optimizer’s intrinsic dynamics may still produce spectral biases that conflict with such external design choices.

\subsection{Spectral Scheduling versus Amplitude Scheduling}

In deep learning optimization, learning-rate and momentum scheduling strategies have long been regarded as important tools for improving model performance. By modulating the training cadence, these methods balance exploration and convergence across different stages, thereby enhancing both convergence speed and generalization ability. Their effectiveness has been repeatedly demonstrated in large-scale training, and they have become a standard component of modern training pipelines.

Mainstream scheduling methods are typically designed as one-dimensional ``amplitude–time'' curves. For example, cosine annealing gradually reduces the learning rate in later training stages to ensure more stable convergence; cyclical learning rates and warm restarts periodically adjust the learning rate to enable transitions between exploration and convergence; the one-cycle policy applies a rise-and-fall pattern within a single training run to dynamically regulate cadence. The common characteristic of these approaches is that they primarily model the variation of update magnitudes over time.

Although amplitude scheduling is effective in practice, it remains limited at the spectral level. Such methods do not directly regulate the relative weighting across frequency bands, and therefore lack the ability to achieve differentiated allocation in the spectral domain. As a result, their influence on spectral behavior is indirect, and they provide no explicit mechanism for broad spectral coverage or cross-band fusion.

\section{First-Order Optimizer Exponent Extension and Preliminary Validation}
\label{sec3}

\subsection{Unified View of SGD and Adam}
\label{sec:unified}

Stochastic Gradient Descent (SGD) with momentum updates parameters as
\begin{equation}
	\label{eq:sgd}
	m_t = \beta_1 m_{t-1} + (1-\beta_1) g_t, \quad
	\theta_{t+1} = \theta_t - \eta\, m_t,
\end{equation}
where $g_t$ is the stochastic gradient, $m_t$ is the momentum term, $\eta$ is the learning rate, and $\beta_1$ controls momentum decay.

Adam extends this by normalizing the momentum term with a second-moment estimate:
\begin{equation}
	\label{eq:adam}
	v_t = \beta_2 v_{t-1} + (1-\beta_2)(g_t \odot g_t), \quad
	\theta_{t+1} = \theta_t - \eta \, \frac{m_t}{v_t^{1/2} + \epsilon},
\end{equation}
where $v_t$ is the element-wise exponential moving average of squared gradients, $\beta_2$ controls its decay, and $\epsilon$ is a small constant to prevent division by zero.
For clarity, Eqs.~\eqref{eq:sgd}--\eqref{eq:adam} are written without bias correction; the unified form below uses the bias-corrected moments.

In practice, both methods often use \emph{bias-corrected} moment estimates
\[
\hat m_t = \frac{m_t}{1-\beta_1^t}, \quad
\hat v_t = \frac{v_t}{1-\beta_2^t},
\]
which remove initialization bias in the early iterations.

These updates can be expressed in the simplified common form
\begin{equation}
	\label{eq:unified}
	\theta_{t+1} = \theta_t - \eta\, \frac{\hat m_t}{\hat v_t^{\,p} + \epsilon},
\end{equation}
where $p \in \mathbb{R}$ determines the influence of the second-moment term.

\noindent
\textbf{Special cases.} Eq.~\eqref{eq:unified} includes several known and extended settings:
\begin{itemize}
	\item $p = 0$: no second-moment normalization (up to a global constant); reduces to momentum SGD when $\beta_1>0$.
	\item $p = 0.5$: Adam/RMSProp-style normalization.
	\item $p \in (0, 0.5)$: intermediate scaling between SGD and Adam; Padam~\cite{chen2021closing} corresponds to a fixed $p$ in this range in its simplified form.
	\item $p \in (0.5, +\infty)$: larger exponents cause proportionally greater reduction in update magnitude for coordinates with large second-moment estimates, emphasizing lower-magnitude directions.
	\item $p \in (-\infty, 0)$: inverse scaling that amplifies updates in coordinates with large second-moment estimates.
\end{itemize}

In the following sections, $p$ will be treated as a continuous control variable. We first analyze its effect on coordinates with varying second-moment magnitudes, and then investigate its role from a sample distribution perspective through both theoretical derivations and empirical studies.

\subsection{Second-Moment Exponent and Gradient Frequency Spectrum}
\label{sec:freq}

The second-moment estimate in Eq.~\eqref{eq:unified} controls the magnitude of updates through
\[
\hat v_t = \frac{v_t}{1-\beta_2^t}, \quad
v_t = \beta_2 v_{t-1} + (1-\beta_2)(g_t \odot g_t),
\]
where $g_t$ denotes the stochastic gradient.
Intuitively, $\hat v_t$ accumulates the history of squared gradients for each coordinate, assigning larger values to directions with persistently large gradients.

\vspace{0.5em}
\noindent\textbf{Fourier-domain interpretation.}
Let the gradient field at iteration $t$ be denoted by $g_t(\mathbf{x})$, where $\mathbf{x}$ indexes spatial or feature-space coordinates.
Its discrete Fourier transform (DFT) is
\[
\mathcal{F}[g_t](\boldsymbol{\omega}) = \sum_{\mathbf{x}} g_t(\mathbf{x})\, e^{-j \boldsymbol{\omega} \cdot \mathbf{x}},
\]
with power spectrum
\[
P_t(\boldsymbol{\omega}) = \bigl|\mathcal{F}[g_t](\boldsymbol{\omega})\bigr|^2, \qquad
V_t(\boldsymbol{\omega}) \triangleq \mathrm{EMA}\big[P_t(\boldsymbol{\omega})\big].
\]
Low frequencies ($\|\boldsymbol{\omega}\|$ small) correspond to smooth, large-scale variations of the gradient (broad, generic patterns).
High frequencies ($\|\boldsymbol{\omega}\|$ large) correspond to rapid changes (sharp boundaries, fine details; in the extreme, noise).

To link coordinate-wise moments to spectral energy, let $H_i(\boldsymbol{\omega})$ denote the effective frequency response of parameter coordinate $i$ (e.g., induced by the model’s filters and computation graph). A Parseval-type relation motivates the approximation
\begin{equation}
	\label{eq:bridge}
	\hat v_{t,i} \;\approx\; \int_{\boldsymbol{\omega}} \big|H_i(\boldsymbol{\omega})\big|^2 \, V_t(\boldsymbol{\omega}) \, d\boldsymbol{\omega},
\end{equation}
which we use as a conceptual bridge between coordinate space and frequency space.

\vspace{0.5em}
\noindent\textbf{Role of $p$ in spectral weighting.}
In the unified update rule
\[
\theta_{t+1} = \theta_t - \eta\, \frac{\hat m_t}{\hat v_t^{\,p} + \epsilon},
\]
the exponent $p$ determines the relative gain applied to gradient components. Aggregating over coordinates via \eqref{eq:bridge}, a frequency-wise view writes
\[
G_p(\boldsymbol{\omega}) \;\propto\; \big[V_t(\boldsymbol{\omega})\big]^{-p}.
\]
When $p > 0$, high-$V_t$ (often high-frequency) components are attenuated, shifting learning toward smoother, low-frequency structures.
When $p < 0$, these components are amplified, emphasizing high-frequency or hard-to-fit patterns.
The special cases $p=0$ (SGD with momentum) and $p=0.5$ (Adam) correspond to uniform (up to a constant) weighting and moderate high-frequency suppression, respectively.

\vspace{0.5em}
\noindent\textbf{Connection to multi-scale methods.}
This spectral interpretation parallels multi-scale optimization strategies: low-$p$ values bias toward fine-scale detail acquisition, while high-$p$ values bias toward coarse-scale, generic pattern learning.
Unlike discrete multi-scale schemes that operate on a few fixed bands, $p$ provides a \emph{continuous}, architecture-agnostic control knob for spectral reweighting within a single formulation.

\emph{Remark.} The frequency-domain view above is a conceptual analogy: the mapping in \eqref{eq:bridge} is heuristic rather than exact, and the actual spectrum depends on both the data distribution and the network architecture.

\subsection{Preliminary Empirical Validation}
\label{sec:preexp}

We conduct controlled experiments to validate the scale-sensitive interpretation of the exponent $p$ proposed in Secs.~\ref{sec:unified}--\ref{sec:freq}.
Experiments are performed on multiple datasets of varying difficulty and noise characteristics: CIFAR-10, CIFAR-100, TinyImageNet, Chaoyang, RAFDB, and FER2013.

All models are trained from scratch under the unified update rule of Eq.~\eqref{eq:unified}.
To isolate the role of the second-moment exponent, we vary $p$ across a wide range including the SGD endpoint ($p{=}0$), the Adam case ($p{=}0.5$), intermediate values, and extrapolated regimes $p \in (-\infty,0)$.
Except for $p$, all training settings are fixed within each run; hyperparameter details are provided in the Appendix.

Evaluation follows standard practice: we report Top-1 test accuracies for each $(\text{architecture}, \text{dataset}, p)$ configuration.
Figure~\ref{R18-pow_test_acc} illustrates representative results for ResNet-18.
The observed trends are consistent with the spectral weighting perspective in Sec.~\ref{sec:freq}:
moderate $p$ values tend to balance low- and high-frequency components, whereas extreme values bias the optimizer toward either strong suppression or amplification of high-frequency content.

\paragraph{Dataset-specific patterns.}
The performance landscape over $p$ varies considerably across datasets.
FER2013, which contains a relatively high proportion of label noise and irregular samples, shows a clear advantage for positive $p$, consistent with the interpretation that larger exponents attenuate high-frequency noise.
By contrast, CIFAR-10 and RAFDB, which have lower noise levels, can benefit from smaller or even negative $p$, allowing stronger emphasis on high-frequency details without substantial accuracy loss.
TinyImageNet, Chaoyang, and CIFAR-100 exhibit more intricate behavior: their curves feature one or more distinct peaks at intermediate $p$ values, suggesting a delicate trade-off between low- and high-frequency information.
Such multi-modal profiles imply that the optimal $p$ for these datasets may depend on the specific frequency composition of the informative signal versus noise.

Overall, these preliminary results support the hypothesis that $p$ acts as a continuous spectral weighting control, with dataset-dependent optima reflecting the underlying distribution of useful features across frequency scales.

\begin{figure}[H]
	\centering
	\begin{tikzpicture}
		\begin{axis}[
			axis y line*=left,
			axis x line*=bottom,
			xlabel={Ciar10},
			ylabel={Test Acc (\%)},
			xmin=-0.32, xmax=0.12,
			width=0.33\textwidth,
			height=4cm
			]
			\addplot[blue, mark=o] table {
				x    y
-0.3	95.425
-0.28	95.349
-0.26	95.413
-0.24	95.372
-0.22	95.506
-0.2	95.44272727
-0.18	95.52181818
-0.16	95.42545455
-0.14	95.40272727
-0.12	95.45454546
-0.1	95.47727273
-0.08	95.46909091
-0.06	95.41090909
-0.04	95.51181818
-0.02	95.52727273
0	95.47909091
0.02	95.36909091
0.04	95.39545455
0.06	95.30454546
0.08	95.30818182
0.1	95.27363636

			};
		\end{axis}
		\end{tikzpicture}
	\begin{tikzpicture}
		\begin{axis}[
		axis y line*=left,
		axis x line*=bottom,
		xlabel={Ciar100},
		xmin=-0.52, xmax=0.22,
		width=0.33\textwidth,
		height=4cm
		]
		\addplot[blue, mark=o] table {
			x    y
			-0.50 77.13
			-0.45 77.66
			-0.40 77.97
			-0.35 77.67
			-0.30 78.32
			-0.25 77.78
			-0.20 77.89
			-0.15 79.52
			-0.10 79.26
			-0.05 78.73
			0.00 78.93
			0.05 78.51
			0.10 76.97
			0.15 76.91
			0.20 77.09
		};
	\end{axis}
		
	\end{tikzpicture}
\begin{tikzpicture}
	\begin{axis}[
		axis y line*=left,
		axis x line*=bottom,
		xlabel={Tiny Imagenet},
		xmin=-0.27, xmax=0.32,
		width=0.33\textwidth,
		height=4cm
		]
		\addplot[blue, mark=o] table {
			x    y
				-0.25	55.56
-0.2	56.67
-0.15	60.55
-0.1	60.07
-0.05	60.18
0	59.84
0.05	58.77
0.1	59.15
0.15	60.07
0.2	60.69
0.25	59.94
0.3	57.44
		};
	\end{axis}
	
\end{tikzpicture}
\begin{tikzpicture}
	\begin{axis}[
		axis y line*=left,
		axis x line*=bottom,
		xlabel={Chaoyang},
		ylabel={Test Acc (\%)},
		xmin=-0.72, xmax=0.52,
		width=0.33\textwidth,
		height=4cm
		]
		\addplot[blue, mark=o] table {
			x    y
				-0.8	72.65
-0.7	76.81
-0.6	80.32
-0.5	80.08
-0.4	81.49
-0.3	80.18
-0.2	76.34
-0.1	74.99
0	80.32
0.1	79.9
0.2	80.08
0.3	81.58
0.4	81.63
0.5	76.25
		};
	\end{axis}
	
\end{tikzpicture}
\begin{tikzpicture}
	\begin{axis}[
		axis y line*=left,
		axis x line*=bottom,
		xlabel={RAFDB},
		xmin=-0.52, xmax=0.5,
		width=0.33\textwidth,
		height=4cm
		]
		\addplot[blue, mark=o] table {
			x    y
-0.5	84.03
-0.475	83.47
-0.45	84.16
-0.425	84.19
-0.4	84.42
-0.375	84.42
-0.35	84
-0.325	84.52
-0.3	84.16
-0.275	83.57
-0.25	83.8
-0.225	84.88
-0.2	84.45
-0.175	85.1
-0.15	84.91
-0.125	84.52
-0.1	84.09
-0.075	84.32
-0.05	83.8
-0.025	83.25
0	84.42
0.025	85.17
0.05	84.29
0.075	84.06
0.1	84.22
0.125	83.64
0.15	83.28
0.175	83.08
0.2	82.89
0.225	82.99
0.25	83.08
0.275	81.98
0.3	82.14
0.325	81.49
0.35	83.34
0.375	82.4
0.4	81.39
0.425	79.47
0.45	77.15
0.475	79.01
		};
	\end{axis}
	
\end{tikzpicture}
\begin{tikzpicture}
	\begin{axis}[
		axis y line*=left,
		axis x line*=bottom,
		xlabel={fer2013},
		xmin=-0.52, xmax=0.5,
		width=0.33\textwidth,
		height=4cm
		]
		\addplot[blue, mark=o] table {
			x    y
		-0.5	53.22
		-0.475	57.09
		-0.45	58.34
		-0.425	61.47
		-0.4	62.64
		-0.375	63.92
		-0.35	63.53
		-0.325	64.81
		-0.3	65.39
		-0.275	66.37
		-0.25	66.15
		-0.225	65.9
		-0.2	67.87
		-0.175	67.21
		-0.15	67.4
		-0.125	67.51
		-0.1	67.93
		-0.075	68.26
		-0.05	67.73
		-0.025	68.6
		0	68.29
		0.025	68.07
		0.05	68.32
		0.075	67.51
		0.1	67.96
		0.125	69.02
		0.15	68.63
		0.175	67.85
		0.2	68.26
		0.225	68.96
		0.25	68.12
		0.275	69.13
		0.3	68.46
		0.325	68.26
		0.35	68.6
		0.375	68.29
		0.4	68.77
		0.425	68.9
		0.45	68.74
		0.475	67.76
		0.5	69.16
		};
	\end{axis}
	
\end{tikzpicture}
	\caption{R18 all: Test ACC vs.\ pow}
	\label{fig:R18-pow_test_acc}
\end{figure}
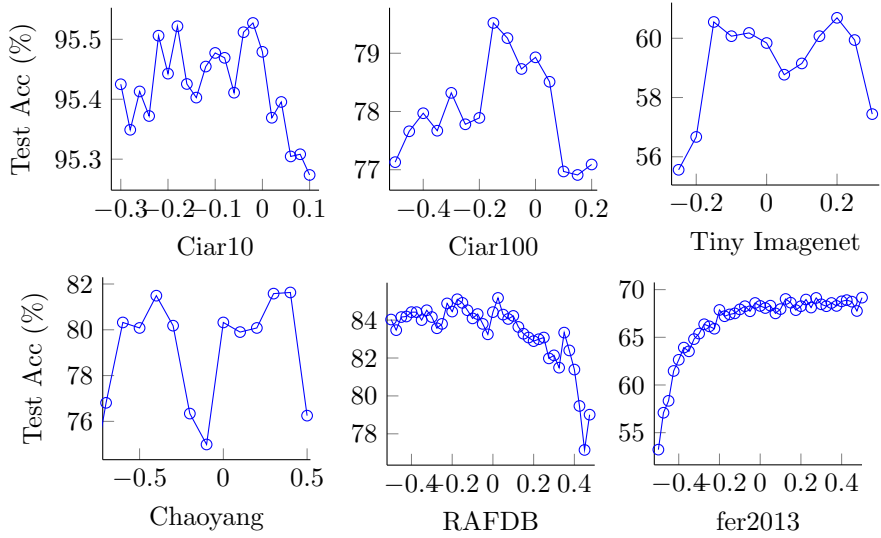

\section{Theoretical Spectral Analysis and Low-Dimensional Demonstrations}
\label{sec4}

\subsection{Spectral Analysis}

\subsubsection{Theoretical Analysis}

We consider a one-dimensional regression with uniform density $\rho\equiv 1$ and focus on how a (possibly time-varying) exponent $p(t)$ shapes spectral decay.
Let $e(x,t)=w(x,t)-y(x)$ be the pointwise error and
\[
A(x,t)\;\triangleq\;\int_0^t e(x,s)^2\,ds
\]
the cumulative local squared error (``energy''). With a locality-modulated step
$\tilde\eta(x,t)=\eta\,(A(x,t)+\varepsilon)^{-p(t)}$ and no momentum, we write
\begin{equation}
	\partial_t e(x,t)\;=\;-\,c_0\,\big(A(x,t)+\varepsilon\big)^{-p(t)}\,e(x,t),
	\qquad
	\partial_t A(x,t)\;=\;e(x,t)^2,
	\label{eq:pointwise_system_new}
\end{equation}
where $c_0$ absorbs constant factors (e.g., $c_0\simeq 2\eta$ in our implementation).

Eliminating $t$ from \eqref{eq:pointwise_system_new} gives
\begin{equation}
	\frac{d\,(e^2)}{dA}\;=\;-\,2c_0\,\big(A+\varepsilon\big)^{-p(t)}.
\end{equation}
For constant $p$ and $p\neq 1$, this integrates to
\begin{equation}
	e(x,t)^2\;=\;e_0(x)^2\;-\;\frac{2c_0}{1-p}\!\left[\big(A(x,t)+\varepsilon\big)^{\,1-p}-\varepsilon^{\,1-p}\right].
	\label{eq:eA_relation_new}
\end{equation}

\paragraph{Cosine scheduler for $p(t)$.}
In our study $p$ varies as a tidal (cosine) schedule
\[
p(t)\;=\;p_{\text{mid}}+p_{\text{amp}}\sin\!\Big(\tfrac{2\pi}{T}\,t+\phi\Big),
\]
with period $T$ and phase $\phi$. It is convenient to introduce an \emph{effective time}
\begin{equation}
	\tau(t)\;\triangleq\;\int_0^t\big(\overline{A}(s)+\varepsilon\big)^{-p(s)}ds,
	\qquad
	\overline{A}(t)=\int_0^1 A(x,t)\,dx,
	\label{eq:tau_time}
\end{equation}
which reduces to $\tau\propto t^{\,1-p}$ under constant $p$ and slowly varying $\overline{A}$.

\paragraph{Fourier expansion and near-diagonality.}
Let $\{\phi_k\}_{k\ge 1}$ be the orthonormal sine basis aligned with our implementation,
\[
\phi_k(x)=\sqrt{2}\,\sin(2\pi k x),\qquad \int_0^1\phi_k(x)^2\,dx=1,
\]
and expand $e(x,t)=\sum_k a_k(t)\phi_k(x)$. Projecting \eqref{eq:pointwise_system_new} yields
\begin{equation}
	\dot a_k(t)\;=\;-\sum_j M_{kj}(t)\,a_j(t),\qquad
	M_{kj}(t)\;=\;c_0\int_0^1\big(A(x,t)+\varepsilon\big)^{-p(t)}\phi_k(x)\phi_j(x)\,dx.
	\label{eq:matrix_coupling}
\end{equation}
When $A$ is nearly spatially uniform (e.g., single/narrow-band energy), $M_{kj}$ is close to diagonal and each mode evolves approximately independently:
\begin{equation}
	\dot a_k(t)\;\approx\;-\,C_k(t)\,a_k(t),\qquad
	C_k(t)\;=\;c_0\int_0^1\big(A(x,t)+\varepsilon\big)^{-p(t)}\phi_k(x)^2\,dx.
	\label{eq:Ck_def}
\end{equation}

\paragraph{Early-time expansion with explicit overlaps.}
For small energy (early time), using
\[
\big(A+\varepsilon\big)^{-p(t)}=\varepsilon^{-p(t)}\Big[1-\tfrac{p(t)}{\varepsilon}A+O(A^2)\Big],
\]
and approximating $A(x,t)\approx \sum_j H_j(t)\,\phi_j(x)^2$ with $H_j(t)=\int_0^t a_j(s)^2 ds$, we obtain
\begin{align}
	C_k(t)&\approx c_0\,\varepsilon^{-p(t)}\!\left[1-\frac{p(t)}{\varepsilon}\sum_j H_j(t)\int_0^1 \phi_k(x)^2\phi_j(x)^2\,dx\right]+O(H^2),\\
	\int_0^1 \phi_k^2\phi_j^2\,dx&=\begin{cases}
		\;\dfrac{3}{2}, & j=k,\\[4pt]
		\;1, & j\neq k,
	\end{cases}
	\label{eq:overlaps}
\end{align}
hence
\begin{equation}
	C_k(t)\;\approx\;c_0\,\varepsilon^{-p(t)}\left[1-\frac{p(t)}{\varepsilon}\Big(\tfrac{3}{2}H_k(t)+\sum_{j\ne k}H_j(t)\Big)\right]+O(H^2).
	\label{eq:Ck_early}
\end{equation}
Equation \eqref{eq:Ck_early} shows that \emph{frequency preference arises from multi-band energy distribution} via the unequal overlaps in \eqref{eq:overlaps}. In the single-band limit ($H_j=0$ for $j\neq k$), $C_k$ is $k$-invariant up to $H_k$, i.e., mere oscillation frequency does not by itself induce a bias.

\paragraph{Time reparameterization and decay law.}
Under the near-diagonal approximation, \(\log|a_k(t)|\approx \log|a_k(0)|-\int_0^t C_k(s)\,ds\).
If we further replace $C_k(s)$ by its spatially averaged surrogate $c_0(\overline{A}(s)+\varepsilon)^{-p(s)}$,
then
\[
\log|a_k(t)|\;\approx\;\log|a_k(0)|-c_0\,\tau(t),
\]
with $\tau(t)$ defined in \eqref{eq:tau_time}. For constant $p\neq 1$, this yields an affine dependence on $t^{\,1-p}$; for $p=1$, it becomes $\log t$.

\paragraph{Remarks.}
(1) The coupling matrix $M_{kj}(t)$ in \eqref{eq:matrix_coupling} quantifies implicit cross-frequency coupling induced by spatial non-uniformity of $\tilde\eta(x,t)$; Eq.~\eqref{eq:Ck_early} provides its early-time footprint.  
(2) With a cosine $p(t)$, the natural independent variable is $\tau(t)$; using $t^{\,1-p_{\mathrm{eff}}}$ is a practical approximation when $p(t)$ varies slowly over the fitting window.  
(3) All continuous integrals are implemented as discrete grid averages in our code; the discrepancy is $O(1/N)$ with $N$ grid points.

\subsubsection{Numerical Verification}
\paragraph{Experimental setup.}
We run a one–dimensional regression on $[0,1)$ under uniform density.  
The target signal is
$y(x)=\sum_{k\in\{1,2,4,8,16,32\}} a_k \sin(2\pi k x)$
with amplitudes $(1.0,\,0.8,\,0.6,\,0.6,\,0.5,\,0.4)$ (broadband but not flat).
We discretize with $N{=}4096$ grid points, time step $dt{=}0.005$, and $T{=}4000$ steps.
The error evolves with a locality–modulated step
$\tilde\eta(x,t)=\eta\,(G(x,t)+\varepsilon)^{-p(t)}$ where $G$ accumulates squared gradients;
we use $\eta{=}0.2$ and $\varepsilon{=}10^{-8}$.
On the orthonormalized $\sin(2\pi kx)$ basis we record the envelope
$|E_k(t)|=\lvert\langle e(\cdot,t),\phi_k\rangle\rvert$ every 10 steps.
To quantify early-time decay we regress
$\log |E_k(t)| \approx \alpha - C_k\,t^{\,1-p}$ on the window $t\in(0,6]$
(for time-varying $p(t)$ we use $p_{\mathrm{eff}}$ in the same form).

We compare two families:
\begin{itemize}
	\item \textbf{Constant $p$:} $p\in\{-0.5,\,-0.25,\,0,\,0.25,\,0.5\}$.
	\item \textbf{Cosine tidal $p(t)$:} $p(t)=p_{\text{mid}}+p_{\text{amp}}\sin(2\pi t/T_{\text{cyc}})$
	with period $T_{\text{cyc}}{=}6.0$ and three ranges
	$0.5\!\leftrightarrow\!-0.5$, $0.5\!\leftrightarrow\!0.0$, and $0.0\!\leftrightarrow\!-0.5$ (phase $\phi{=}0$).
\end{itemize}

\paragraph{Design rationale.}
(i) A broadband set of $k$’s probes frequency preference and cross-mode coupling.  
(ii) The early window $t\le 6$ limits late-time nonlinear mixing so that $|E_k|$ is closer to piecewise-exponential.  
(iii) The $p$ grid spans both signs; the cosine scheduler tests time-variation (same mean, higher variance) against the constant-$p$ baselines.  
(iv) $\varepsilon$ prevents degeneracy as $G\!\to\!0$; $\eta$ is chosen so all settings enter a stable fitting regime within the window.

\begin{figure}[t]
	\centering
	\includegraphics[width=0.78\linewidth]{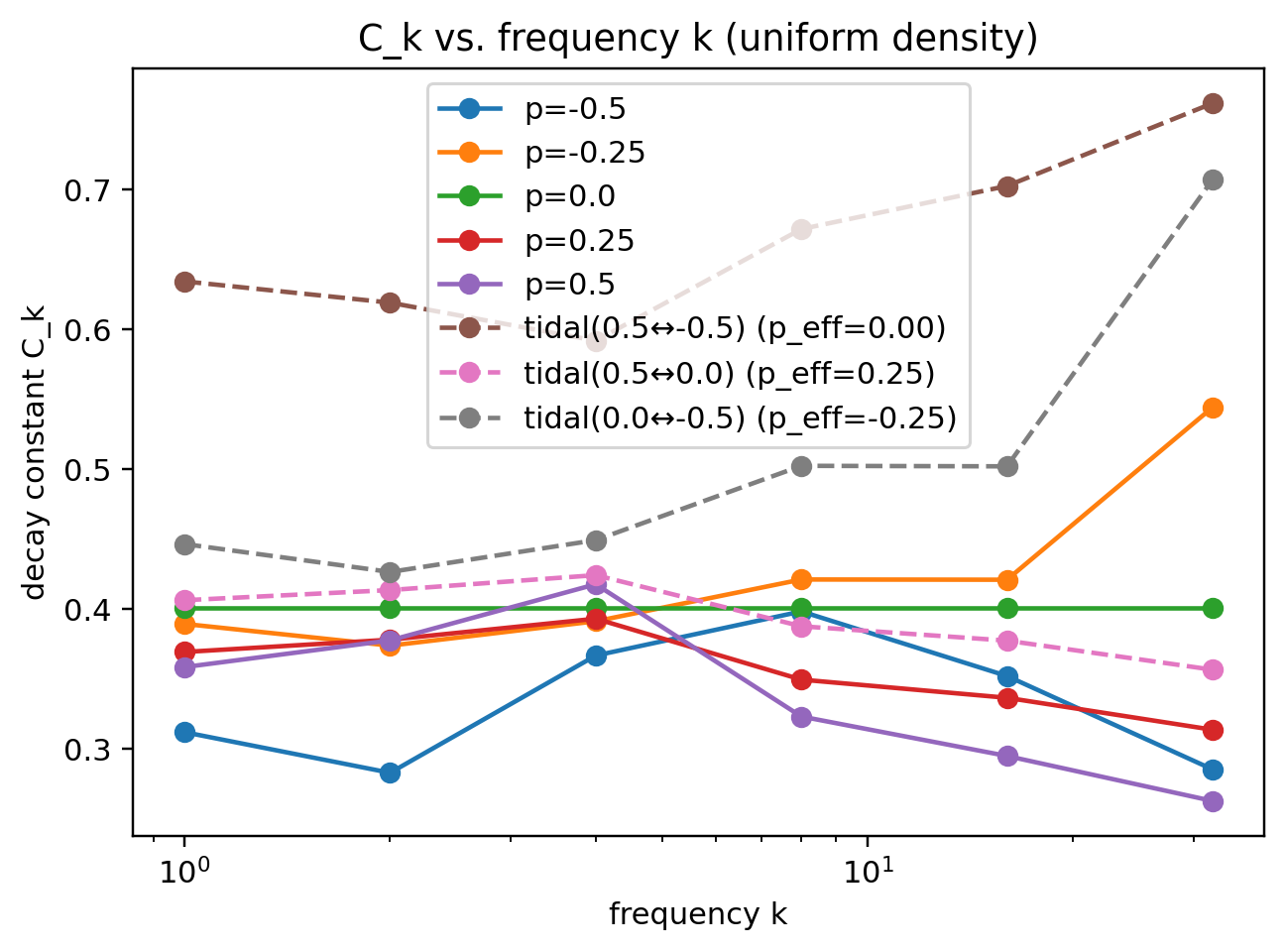}
	\caption{$C_k$ vs.\ frequency $k$ (uniform density). Constant $p$ and three cosine tidal ranges are overlaid.}
	\label{fig:ckvk}
\end{figure}

\begin{figure}[t]
	\centering
	\includegraphics[width=0.95\linewidth]{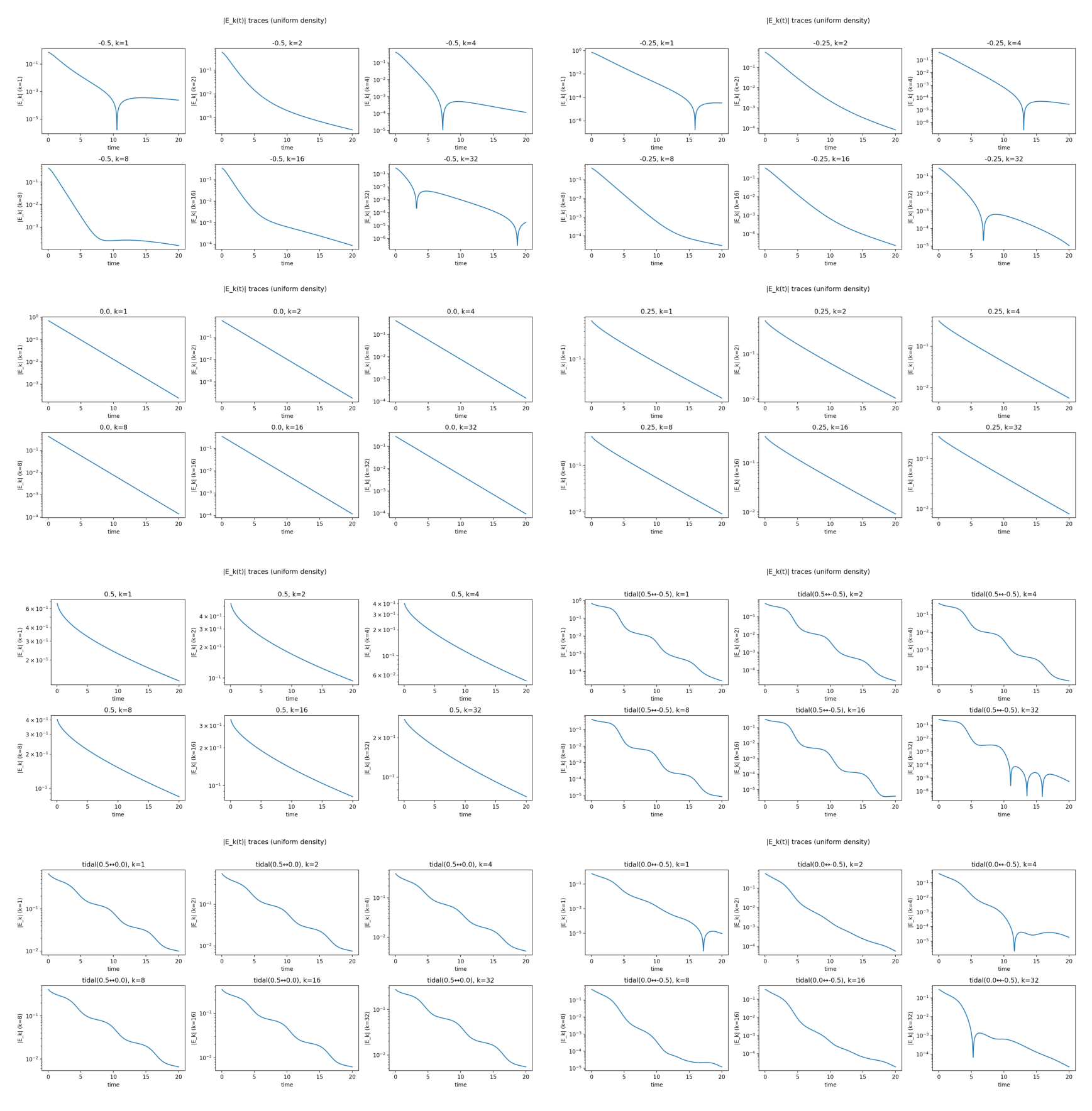}
	\caption{Mosaic of $|E_k(t)|$ traces for all eight configurations ($2{\times}4$).}
	\label{fig:ekmosaic}
\end{figure}

\paragraph{Summary figures.}
Fig.~\ref{fig:ckvk} shows $C_k$ vs.\ frequency $k$ (log $x$-axis) for all settings.
Fig.~\ref{fig:ekmosaic} is a $2{\times}4$ mosaic of the eight $|E_k(t)|$ panels
(5 constant $p$’s + 3 tidal schedules), all with log $y$-axis.

\paragraph{Quantitative results.}
Early-time decay constants ($C_k$, larger is faster) and fit quality ($R^2$) are:

\begin{center}
	\footnotesize
	\begin{tabular}{lrrrrrr}
		\toprule
		$p$/key & C(1)&C(2)&C(4)&C(8)&C(16)&C(32)\\
		\midrule
		$-0.5$ & 0.312&0.283&0.366&0.398&0.352&0.285\\
		$-0.25$& 0.389&0.374&0.391&0.421&0.421&0.544\\
		$0.0$  & 0.400&0.400&0.400&0.400&0.400&0.400\\
		$0.25$ & 0.369&0.378&0.393&0.349&0.336&0.313\\
		$0.5$  & 0.358&0.377&0.417&0.323&0.295&0.263\\
		\midrule
		tidal $(0.5\!\leftrightarrow\!-0.5)$ ($p_{\text{eff}}{=}0.00$)
		&0.634&0.619&0.592&0.672&0.703&0.762\\
		tidal $(0.5\!\leftrightarrow\!0.0)$ ($p_{\text{eff}}{=}0.25$)
		&0.406&0.413&0.424&0.387&0.377&0.356\\
		tidal $(0.0\!\leftrightarrow\!-0.5)$ ($p_{\text{eff}}{=}-0.25$)
		&0.446&0.426&0.449&0.502&0.502&0.707\\
		\bottomrule
	\end{tabular}
\end{center}

\paragraph{Observations and analysis.}
\begin{enumerate}
	\item \textbf{Constant $p$: clean frequency preference.}
	For $p{=}0$ all modes decay at the same rate ($C_k\equiv 0.4$), matching a global step size; the traces are straight lines on a log scale.
	Positive $p$ exhibits a mild \emph{low-pass} bias (smaller $C_k$ at higher $k$), while negative $p$ shows a \emph{high-pass} tendency (larger $C_k$ at high $k$).
	\item \textbf{Strongly negative $p$ induces non-single-exponential behavior.}
	For $p{=}-0.5$, $R^2$ drops at high $k$ (e.g., $0.55$ at $k{=}32$) and the traces display dips/ rebounds, indicating spatially non-uniform $\tilde\eta(x,t)$ and implicit cross-frequency coupling.
	\item \textbf{Cosine tidal schedules accelerate early decay.}
	With similar effective mean $p_{\mathrm{eff}}$, time variation increases $\mathbb{E}[\tilde\eta]$ in the early regime ($G{+}\varepsilon<1$), yielding larger $C_k$ than the corresponding constant-$p$ baselines.
	The widest swing, tidal $(0.5\!\leftrightarrow\!-0.5)$, outperforms $p{=}0$ by a large margin across all $k$ (e.g., $C_{32}{=}0.762$ vs.\ $0.400$), but exhibits multi-slope segments and slightly lower $R^2$.
	\item \textbf{Intermediate swings behave as expected.}
	Tidal $(0.5\!\leftrightarrow\!0.0)$ ($p_{\mathrm{eff}}{=}0.25$) remains low-pass and only modestly faster than constant $p{=}0.25$;
	tidal $(0.0\!\leftrightarrow\!-0.5)$ ($p_{\mathrm{eff}}{=}-0.25$) notably accelerates high frequencies (e.g., $C_{32}$ from $0.544$ to $0.707$) while keeping high $R^2$.
\end{enumerate}

\paragraph{Takeaways.}
(i) The sign of $p$ sets the low-/high-pass direction; constant $p$ gives the cleanest pattern.  
(ii) Cosine tidal $p(t)$ delivers substantial early-time acceleration (larger $C_k$), particularly at high $k$, at the cost of multi-slope behavior (slightly reduced $R^2$).  
(iii) Strongly negative $p$ magnifies spatial non-uniformity and implicit coupling, producing non-single-exponential traces (dips and rebounds), which motivates using an effective time scale in later sections.

\subsection{Low-Dimensional Demonstrations}
\subsubsection{One-Dimensional Demonstration}
We consider a stylized one-dimensional regression setting to isolate the effect of the second-moment exponent $p$.
Let the target be $y(x)=x$ for $x\in[0,1]$ and the initial predictor $w(x,0)=0$, so that the initial error is
\[
e(x,0)=w(x,0)-x=-x.
\]
We use a weighted quadratic loss with a simple, positive sample density
\[
L(t)=\int_0^1 \bigl[w(x,t)-x\bigr]^2\,\rho(x)\,dx,
\qquad
\rho(x)=m\,(x-0.5)+0.5,\;\; m\in(-1,1).
\]
The functional gradient is
\[
\frac{\delta L}{\delta w(x,t)}=2\,\rho(x)\,e(x,t),
\]
and we write the local (per-$x$) gradient as $g(x,t)=2\,\rho(x)\,e(x,t)$.

\label{sec:early}
Define the accumulated squared gradient
\[
G(x,t)=\int_0^t g(x,s)^2\,ds.
\]
Under an early-time approximation (i.e., $e(x,s)\approx e(x,0)=-x$ for $s\in[0,t]$), we have
\[
G(x,t)\approx \int_0^t \bigl[2\,\rho(x)\,(-x)\bigr]^2 ds
=4\,x^2\,\rho(x)^2\,t.
\]
We model the unified update (local learning-rate modulation) as
\[
\tilde{\eta}(x,t)=\eta\,\bigl[G(x,t)+\epsilon\bigr]^{-p},
\]
which yields the error dynamics
\[
\frac{d e(x,t)}{dt}
=-\,2\,\tilde{\eta}(x,t)\,\rho(x)\,e(x,t)
\approx
-\,2\,\eta\,\rho(x)\,\bigl(4\,x^2\,\rho(x)^2\bigr)^{-p}\,t^{-p}\,e(x,t).
\]
Separating variables,
\[
\frac{d}{dt}\ln|e(x,t)|
=-\,2\,\eta\,\rho(x)\,\bigl(4\,x^2\,\rho(x)^2\bigr)^{-p}\,t^{-p}.
\]
Integrating from $0$ to $t$ (for $p\neq 1$) gives
\[
\ln\frac{|e(x,t)|}{|e(x,0)|}
= -\,\frac{2\eta}{1-p}\,4^{-p}\,x^{-2p}\,\rho(x)^{\,1-2p}\,t^{\,1-p}.
\]
Using $|e(x,0)|=x$, we obtain the closed-form early-time decay
\begin{equation}
	\label{eq:1d_decay}
	|e(x,t)|=x\,\exp\!\Bigl[-\,F(x;m,p,\eta)\,t^{\,1-p}\Bigr],
	\quad
	F(x;m,p,\eta)=\frac{2\eta}{1-p}\,4^{-p}\,x^{-2p}\,\rho(x)^{\,1-2p}.
\end{equation}

\paragraph{Remarks.}
(i) The exponent $1-p$ controls the time scale of decay, linking $p$ to the effective learning schedule.
(ii) The spatial factor $x^{-2p}\rho(x)^{1-2p}$ shows how $p$ interacts with the sample density $\rho$ to reweight local decay rates.

\textbf{Adam/RMSProp regime ($p=0.5$).}
Since $1-2p=0$, the density term is screened out:
\[
F(x;m,0.5,\eta)=\frac{2\eta}{1-0.5}\,4^{-0.5}\,x^{-1}
=\frac{2\eta}{x}.
\]
Hence, in the early-time limit, the dependence on $m$ disappears from \eqref{eq:1d_decay}, matching the empirical observation that $p=0.5$ reduces sensitivity to sample density.

\medskip
\noindent
\textbf{SGD endpoint ($p=0$).}
We have $F(x;m,0,\eta)=2\eta\,\rho(x)$, so the decay rate scales linearly with the local density $\rho(x)$.

\medskip
\noindent
\textbf{Negative exponents ($p\in(-\infty,0)$).}
The factor $x^{-2p}\rho(x)^{\,1-2p}$ amplifies directions with larger initial gradient magnitude (higher $x$ and/or larger $\rho$), predicting earlier emphasis on high-variation regions.

To complement the early-time analysis, we run a simple one-dimensional numerical iteration 
following the unified update rule in \eqref{eq:1d_decay}, but without any exponential moving averages or momentum terms.
The goal is not to tune an optimizer, but to directly visualize how different values of $p$ 
interact with the slope parameter $m$ in the sample density
\[
\rho(x) = m\,(x-0.5) + 0.5,
\]
where $m$ controls the density gradient from the left boundary ($x=0$) to the right boundary ($x=1$):
\begin{itemize}
	\item $m>0$: higher sample density near $x=1$;
	\item $m<0$: higher sample density near $x=0$;
	\item $m=0$: uniform density.
\end{itemize}

In each experiment, we initialize $w(x)=0$ on $x\in[0,1]$, 
and iterate the discrete form of \eqref{eq:1d_decay} for $T=3000$ steps 
with a fixed learning rate $\eta$ and small $\epsilon$ for numerical stability.
We track $w(x)$ at several checkpoints $t\in\{200,500,1000,2000,3000\}$,
and plot it against the target $y(x)=x$.
To quantify accuracy at the final step, we report the fractional root-mean-squared error (fRMSE)
on the boundary region $x\in[0.8,1]$.

\begin{figure}[H]
	\centering
	\includegraphics[width=0.9\textwidth]{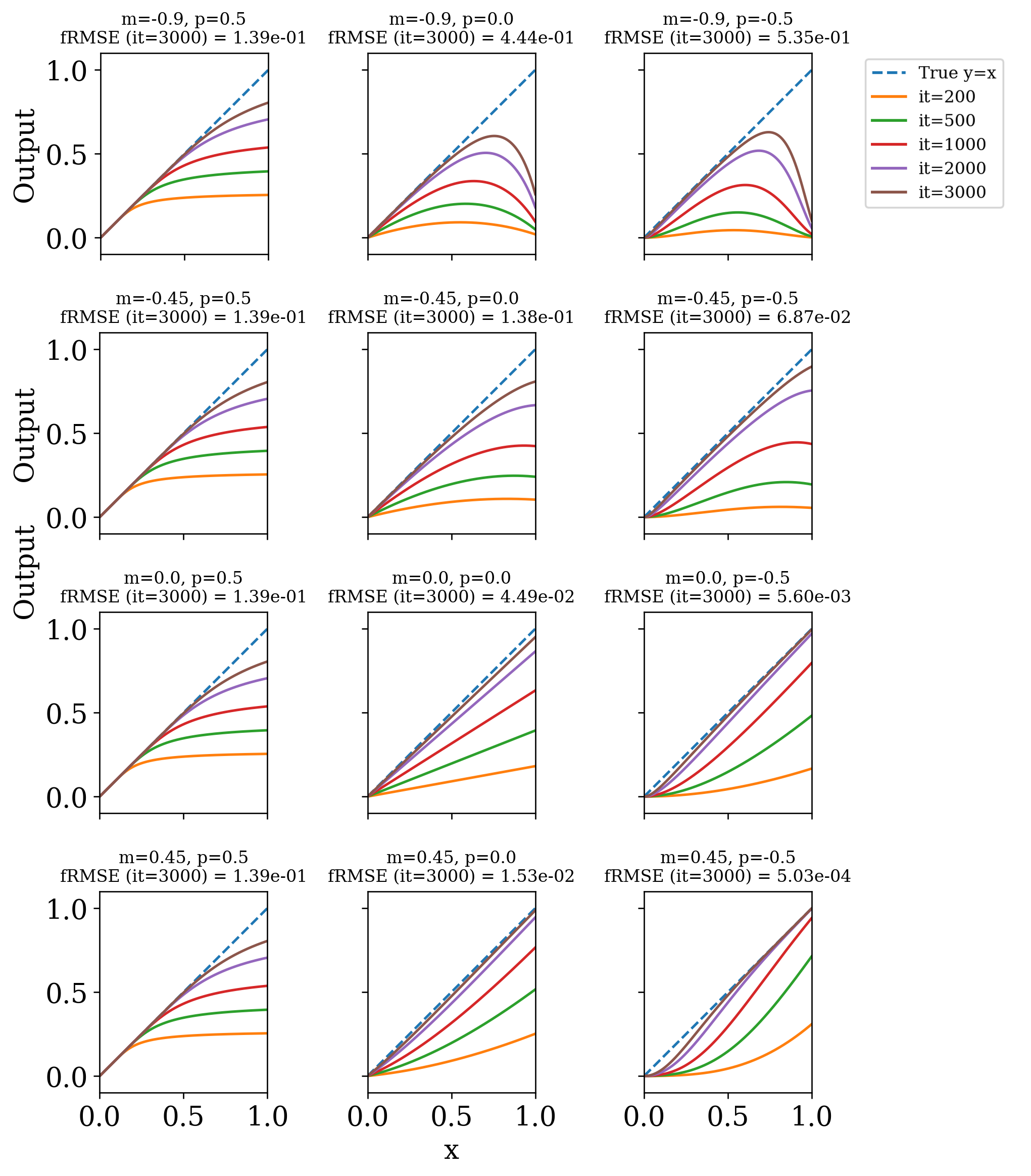}
	\caption{
		Discrete-time 1D demonstrations for different density slopes $m$ (rows) and second-moment exponents $p$ (columns).
		Each panel shows the target $y=x$ (dashed), predictions at multiple iterations (solid), 
		and the final-step fRMSE on the boundary $x\in[0.8,1]$.
		Trends are consistent with the qualitative consequences in Sec.~\ref{sec:special}.
	}
	\label{fig:combined_progress}
\end{figure}

Figure~\ref{fig:combined_progress} shows the results for four representative $m$ values 
and three representative $p$ values ($0.5$, $0$, $-0.5$).
The dashed line is the ground truth $y=x$,
solid curves show the predictions at different iterations,
and the fRMSE is displayed for $t=3000$ only.
The plots illustrate the qualitative trends predicted by Sec.~\ref{sec:early}:
\begin{itemize}
	\item For $p=0.5$ (Adam/RMSProp regime), the curves are nearly insensitive to $m$.
	\item For $p=0$ (SGD endpoint), higher $m$ accelerates convergence in the high-density boundary.
	\item For $p<0$, this density sensitivity is amplified: alignment between $m$ and the evaluation region 
	yields much faster decay, while misalignment slows it down.
\end{itemize}

\subsubsection{Two-Dimensional Demonstration}
Despite its simplicity, the 1D model produces clear, testable scaling trends in how $p$ interacts with the sample density slope $m$, 
but it is confined to an axis-aligned geometry without spatial correlation. 
To examine whether the $p$-dependent early emphasis persists in a richer spatial setting, we next construct a controlled two-dimensional classification problem with known density asymmetry and localized difficulty, 
bridging the gap between analytically tractable 1D cases and the complexity of real-world data.

To further examine whether these observations persist beyond the toy setting, 
we next construct a controlled two-dimensional synthetic dataset. 
This allows us to introduce spatial structure, variable density, and localized difficulty in a setting where all ground-truth factors are known, 
bridging the gap between the tractable 1D derivation and high-dimensional real-world tasks.

To further verify the qualitative trends predicted by the one-dimensional analysis, 
we construct a controlled two-dimensional binary classification task.
The dataset is manually designed to contain two intersecting ``angle''-shaped class manifolds
with asymmetric sampling density: for each class, $75\%$ of the points lie in a ``far'' branch
and $25\%$ in a ``near'' branch. This setup intentionally produces 
regions of different gradient magnitudes at initialization, mimicking heterogeneous local difficulty.

We train a small two-layer MLP (\(20\) hidden units, ReLU activation) under the unified update rule
with various second-moment exponents \(p\in\{-0.10,-0.05,0,0.25,0.50\}\), 
using the same learning rate and without weight decay. 
All runs start from identical random seeds for fair visual comparison.
For each configuration, we record decision boundaries at multiple training checkpoints
(early to late iterations) and horizontally concatenate them into a single row,
so that columns correspond to training iterations and rows to different \(p\) values.

\begin{figure}[H]
	\centering
	\includegraphics[width=\linewidth]{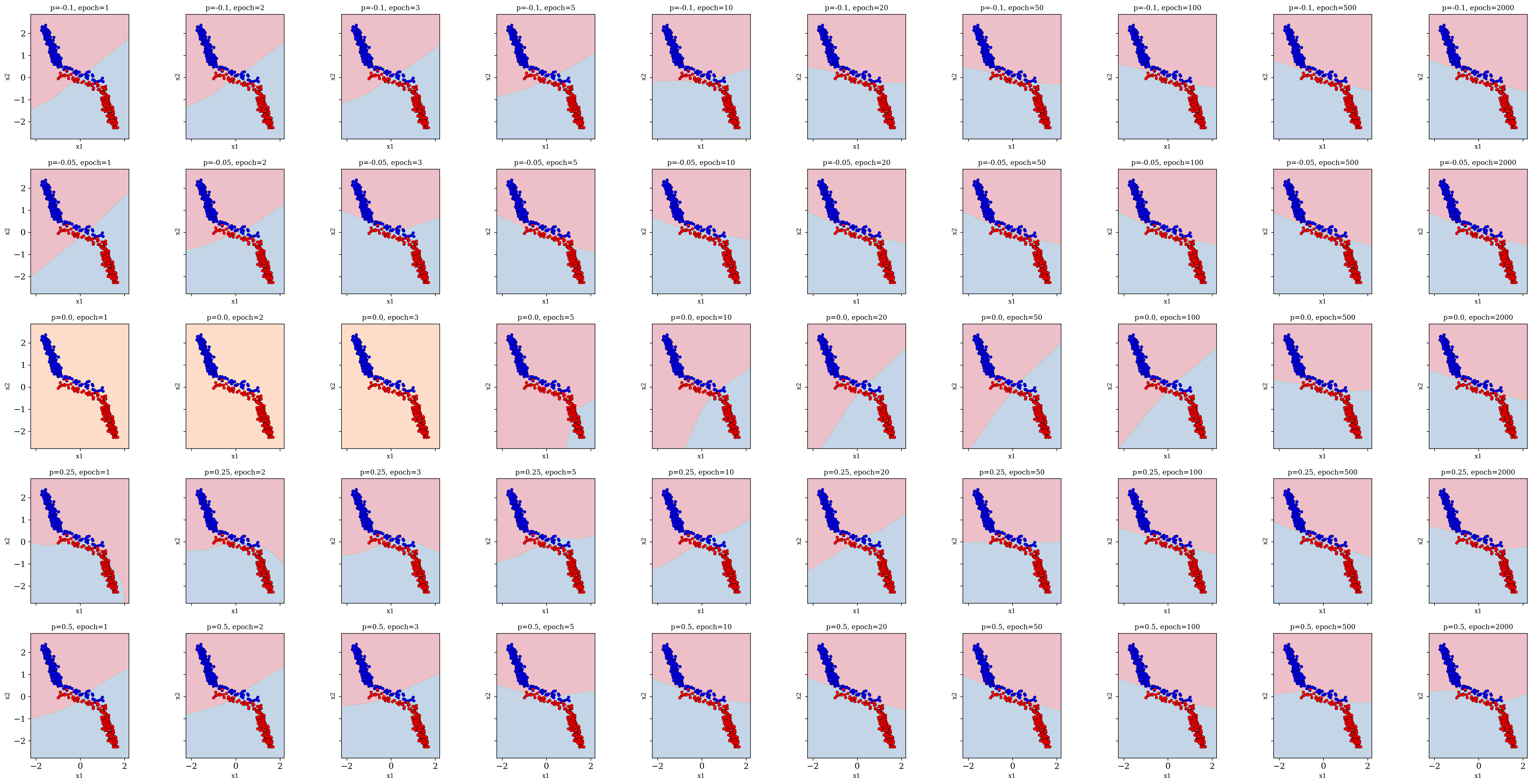}
	\caption{
		Decision boundary evolution for different $p$ values on the synthetic 2D dataset.
		Rows: different $p$; Columns: increasing training iterations from left to right.
		High effective-update samples (top 30\% per class) are marked with colored borders.
	}
	\label{fig:biggrid_p_by_epoch}
\end{figure}

Figure~\ref{fig:biggrid_p_by_epoch} shows the resulting decision boundary evolution.
We highlight samples in the top-$30\%$ of per-class effective update norms with a distinct border color,
providing an intuitive view of where the optimizer allocates larger early updates.

\paragraph{Observations.}
\begin{itemize}
	\item The SGD endpoint (\(p=0\)) is the slowest to align its decision boundary with the optimal separator; significant misclassification persists even after several hundred iterations.
	\item Slightly negative exponents (\(p=-0.05\)) produce the fastest early alignment, achieving near-correct separation within the first few iterations.
	\item \(p=-0.10\) and \(p=0.50\) both show competitive early performance, though slightly behind \(p=-0.05\) in the sharpness of initial boundary rotation.
	\item Positive \(p=0.25\) is noticeably slower than \(p=-0.05\) and \(p=0.50\), suggesting that moderate positive exponents can dampen early correction of the difficult regions.
\end{itemize}

\paragraph{Interpretation.}
These results are consistent with the one-dimensional prediction that negative \(p\) values
emphasize high-gradient regions early, accelerating the alignment of the decision boundary.
The best-performing \(p\) in this synthetic setup is \(-0.05\), but we stress that 
the optimal exponent is dataset-dependent, and different geometries or difficulty distributions 
may shift this optimum.

\section{p-Tidal Scheduling}
\label{sec5}

We evaluate the proposed $p$-tidal scheduling strategy through 
separate experiments on different backbone architectures. 
Each subsection presents one experiment, including the setup, 
results, and discussion.

\subsection{ResNet-18 on TinyImageNet}

\subsubsection{Experimental Setup}
We train ResNet-18 on the TinyImageNet dataset 
(200 classes, $64 \times 64$ images, 100k training and 10k validation samples).
Standard data augmentations (random cropping and horizontal flipping) are applied. 
Training runs for 200 epochs with a batch size of 256. 

The optimizer is SGD with momentum 0.9, learning rate $0.1$, 
and weight decay $5 \times 10^{-4}$. 
\begin{itemize}
	\item \textbf{Baseline:} fixed $p=0$, corresponding to the SGD endpoint. 
	\item \textbf{Proposed:} pulse–tidal scheduling with 
	$p=-0.05$ in the pulse phase, and tidal values alternating 
	between $0.25$ and $-0.15$ every 200 iterations. 
\end{itemize}

\subsubsection{Results}
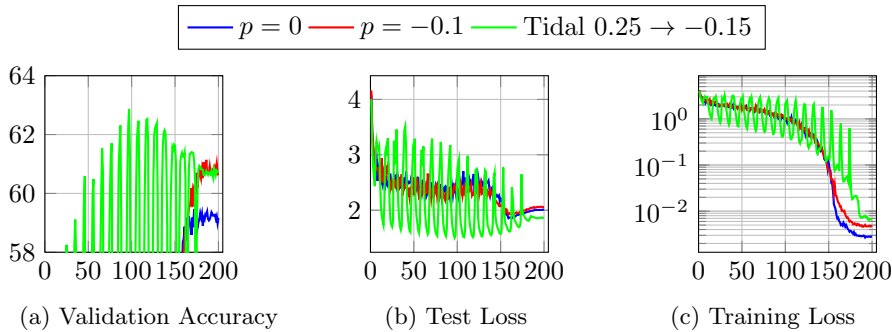
\begin{figure}[H]
	\centering
	\begin{subfigure}[b]{0.3\textwidth}
		\centering
		\begin{tikzpicture}
			\begin{axis}[
				width=\textwidth,
				height=\textwidth,
				ymin=58, ymax=64,
				xmin=0, xmax=205,
				legend style={at={(2.4,1.4)}, anchor=north,legend columns=3},
				grid=both,
				]
				\addplot[thick,blue] table {
					x    y
					1	11.56000042
					2	22.77000046
					3	27.52000046
					4	30.55999947
					5	33.63999939
					6	34.24000168
					7	32.86999893
					8	39.66999817
					9	35.75
					10	37.75
					11	34.40999985
					12	38.84000015
					13	40.88000107
					14	41.59999847
					15	38.50999832
					16	41.72999954
					17	42.81000137
					18	39.16999817
					19	41.83000183
					20	43.15000153
					21	41.72999954
					22	41
					23	40.36000061
					24	42.93000031
					25	42.38000107
					26	43.36000061
					27	40.00999832
					28	40.33000183
					29	40.88000107
					30	42.79000092
					31	39.16999817
					32	42.70999908
					33	43
					34	42.22999954
					35	44.38000107
					36	44.20000076
					37	44.43000031
					38	44.25999832
					39	41.33000183
					40	43.31999969
					41	38.72000122
					42	43.68999863
					43	42.02999878
					44	42.70000076
					45	44.41999817
					46	43.52000046
					47	41.93999863
					48	43.27000046
					49	43.49000168
					50	44.18999863
					51	43.06999969
					52	45.22000122
					53	43.25999832
					54	42.56999969
					55	41.18999863
					56	46.54999924
					57	45.68000031
					58	45.18000031
					59	46.75
					60	43.15000153
					61	43.08000183
					62	43.18999863
					63	45.59000015
					64	48.34999847
					65	46.65000153
					66	44.93999863
					67	41.61000061
					68	45.22999954
					69	45.52000046
					70	47.06000137
					71	44.79000092
					72	46.52999878
					73	45.56999969
					74	45.99000168
					75	45.52999878
					76	46.09000015
					77	45.45999908
					78	46.04000092
					79	47.65000153
					80	44.90999985
					81	45.90000153
					82	48.18000031
					83	46.77999878
					84	47.09000015
					85	45.95000076
					86	46.40999985
					87	48.34999847
					88	45.47000122
					89	47.38000107
					90	46.61000061
					91	47.25
					92	49.50999832
					93	48.22999954
					94	46.02999878
					95	49.04999924
					96	45.99000168
					97	46.36000061
					98	48.50999832
					99	45.68000031
					100	47.88999939
					101	48.06999969
					102	49.95999908
					103	46.84999847
					104	46.54000092
					105	45.79000092
					106	46.45999908
					107	44.27000046
					108	48.18999863
					109	46.11000061
					110	48
					111	46.68999863
					112	45.95000076
					113	46.61000061
					114	46.02999878
					115	47.34999847
					116	48.18000031
					117	47.56000137
					118	47.41999817
					119	45.75
					120	49.75
					121	47.45000076
					122	47.97000122
					123	50.11000061
					124	46.93000031
					125	48.04999924
					126	49.61999893
					127	49.38999939
					128	47.09000015
					129	48.18000031
					130	49.56000137
					131	47.40000153
					132	46.90999985
					133	47.25
					134	48.31999969
					135	48.88000107
					136	48.65000153
					137	49.40000153
					138	52.00999832
					139	49.90999985
					140	49.27999878
					141	51.13999939
					142	50.45000076
					143	50.11000061
					144	51.13999939
					145	52.77999878
					146	51.77999878
					147	52.38999939
					148	51.77999878
					149	53.88999939
					150	52.66999817
					151	52.86999893
					152	53.13000107
					153	54.72999954
					154	54.68999863
					155	55.61000061
					156	57.06999969
					157	57.34999847
					158	57.25
					159	58.36000061
					160	58.45999908
					161	58.86999893
					162	58.38000107
					163	59.16999817
					164	58.86000061
					165	58.59000015
					166	59.22999954
					167	59.00999832
					168	58.93999863
					169	58.59000015
					170	59.18000031
					171	58.84999847
					172	59.47000122
					173	59.02999878
					174	59.29000092
					175	59.18999863
					176	58.84999847
					177	59.02000046
					178	59.13000107
					179	59.33000183
					180	59.18000031
					181	59.31999969
					182	58.90000153
					183	59.16999817
					184	59.11000061
					185	59.29999924
					186	59.27000046
					187	59.38000107
					188	59.13000107
					189	59.08000183
					190	59.11999893
					191	59.22000122
					192	59.36000061
					193	59.29999924
					194	59.24000168
					195	59.25
					196	59.27000046
					197	59.13000107
					198	59.22999954
					199	58.99000168
					200	59.16999817
				};
				\addlegendentry{$p=0$}
				
				\addplot[thick,red] table {
					x    y
					1	10.64999962
					2	20.75
					3	26.18000031
					4	30.06999969
					5	31.19000053
					6	33.63999939
					7	35.52999878
					8	35.25999832
					9	35.09000015
					10	39.75999832
					11	39.95000076
					12	37.84000015
					13	33.52999878
					14	40.08000183
					15	36.90000153
					16	40.27000046
					17	40.72000122
					18	37.75999832
					19	37.90000153
					20	36.54000092
					21	39.97000122
					22	41.52000046
					23	42.72000122
					24	41.18000031
					25	42.22000122
					26	41.38000107
					27	40.79000092
					28	42.75999832
					29	43.49000168
					30	41.54000092
					31	41.59999847
					32	44.52999878
					33	40.88000107
					34	43.61000061
					35	40.86999893
					36	41.06999969
					37	46.50999832
					38	45.29000092
					39	44.59999847
					40	42.34000015
					41	42.45999908
					42	43.33000183
					43	45.52999878
					44	44.31999969
					45	47.22999954
					46	44.34999847
					47	43.63999939
					48	42.74000168
					49	41.15999985
					50	46.84000015
					51	42.43000031
					52	42.59999847
					53	42.41999817
					54	42.43999863
					55	45.38999939
					56	47.11000061
					57	47.75
					58	44.83000183
					59	43.27000046
					60	46.61000061
					61	44.93999863
					62	45.84000015
					63	46.75
					64	44.58000183
					65	46.84999847
					66	41.59000015
					67	43.5
					68	46.04999924
					69	44.95000076
					70	48.13999939
					71	45.65999985
					72	46.15000153
					73	47.88999939
					74	45.15999985
					75	46.15999985
					76	44.38000107
					77	49.02000046
					78	46.5
					79	46.20000076
					80	47.81999969
					81	47.47999954
					82	45.15000153
					83	48.04999924
					84	47.20999908
					85	48.50999832
					86	49.40000153
					87	46.97000122
					88	46.02999878
					89	49.08000183
					90	45.70000076
					91	46.56999969
					92	48.83000183
					93	47.29000092
					94	48.31999969
					95	48.27000046
					96	45.16999817
					97	45.63999939
					98	46.84000015
					99	47.00999832
					100	47.79000092
					101	48.34999847
					102	48.45000076
					103	48.97000122
					104	47.43000031
					105	47.06000137
					106	48.93999863
					107	47.49000168
					108	47.13999939
					109	48.09999847
					110	47.61999893
					111	47.56999969
					112	47.24000168
					113	47.72999954
					114	48.09999847
					115	46.63999939
					116	45.70000076
					117	49.41999817
					118	48.20000076
					119	49.63999939
					120	49.58000183
					121	48.40999985
					122	48.16999817
					123	49.38999939
					124	47.81999969
					125	50.47000122
					126	48.75999832
					127	46.70000076
					128	50.75999832
					129	49.88999939
					130	48.18999863
					131	48.15999985
					132	49.11000061
					133	48.33000183
					134	50.45000076
					135	49.34999847
					136	49.61999893
					137	50.54999924
					138	49.5
					139	50.52999878
					140	50.18999863
					141	51.11000061
					142	50.11000061
					143	50.79999924
					144	51.84999847
					145	53.22999954
					146	53.22000122
					147	52.36999893
					148	53.63999939
					149	54.61999893
					150	53.83000183
					151	53.5
					152	53.84000015
					153	54.88999939
					154	55.36000061
					155	55.63999939
					156	56.61000061
					157	56.47999954
					158	57.59999847
					159	57.68999863
					160	58.02000046
					161	58.41999817
					162	58.74000168
					163	59.06999969
					164	58.79999924
					165	59.22000122
					166	59.50999832
					167	59.56999969
					168	59.77000046
					169	60.18000031
					170	59.54999924
					171	60.22000122
					172	60.15000153
					173	60.18999863
					174	60.11000061
					175	60.70000076
					176	60.20000076
					177	60.54000092
					178	60.77999878
					179	60.68000031
					180	60.50999832
					181	60.79999924
					182	60.52000046
					183	61.06000137
					184	60.68000031
					185	60.70999908
					186	61.13999939
					187	60.68999863
					188	60.66999817
					189	60.68000031
					190	60.91999817
					191	60.68999863
					192	60.74000168
					193	60.65999985
					194	60.83000183
					195	60.90000153
					196	60.74000168
					197	60.79999924
					198	61.00999832
					199	60.65999985
					200	60.61999893

				};
				\addlegendentry{$p=-0.1$}
				
				\addplot[thick,green] table {
					x    y
					1	12.63000011
					2	25.98999977
					3	32.84999847
					4	38.02000046
					5	41.13000107
					6	29.87999916
					7	26.46999931
					8	29.29000092
					9	29.34000015
					10	27.79000092
					11	39.63000107
					12	48.95999908
					13	53.20000076
					14	55.68999863
					15	56.86000061
					16	48.56999969
					17	32.16999817
					18	30.62000084
					19	31.69000053
					20	27.79000092
					21	36.38999939
					22	49.16999817
					23	55.47000122
					24	57.52999878
					25	58.25
					26	55.11999893
					27	31.40999985
					28	34.93000031
					29	33.29000092
					30	27.52000046
					31	37.72000122
					32	48.86000061
					33	54.66999817
					34	58.22000122
					35	59.11000061
					36	57.93000031
					37	41.97999954
					38	34.91999817
					39	27.09000015
					40	27.39999962
					41	29.28000069
					42	48.50999832
					43	54.18000031
					44	57.43999863
					45	59.63999939
					46	60.59000015
					47	48.66999817
					48	37.97000122
					49	33.79000092
					50	34.24000168
					51	29.12999916
					52	48.41999817
					53	53.83000183
					54	56.95000076
					55	59.93999863
					56	60.49000168
					57	55.90000153
					58	36.45000076
					59	32.22999954
					60	35.33000183
					61	30.54000092
					62	45.40000153
					63	52.83000183
					64	57.90000153
					65	60.15000153
					66	61.52999878
					67	60.61000061
					68	47.04999924
					69	38.65000153
					70	33.77000046
					71	30.14999962
					72	39.65999985
					73	53.54000092
					74	58
					75	59.99000168
					76	61.29000092
					77	61.70999908
					78	53.90999985
					79	40.74000168
					80	35.70999908
					81	31.25
					82	33.93000031
					83	51.04999924
					84	57.34000015
					85	60.79999924
					86	61.63000107
					87	62.31000137
					88	59.13000107
					89	44.31000137
					90	39.34999847
					91	34.77999878
					92	31.12999916
					93	50.04999924
					94	57.29999924
					95	60.54000092
					96	62.18999863
					97	62.86999893
					98	61.38000107
					99	53.31999969
					100	37.77999878
					101	37.61999893
					102	35.45000076
					103	51.20999908
					104	57.02000046
					105	59.81000137
					106	61.79000092
					107	62.24000168
					108	62.47000122
					109	59.95999908
					110	46.47999954
					111	40.84999847
					112	36.22999954
					113	48.20999908
					114	56.29999924
					115	59.84999847
					116	61.74000168
					117	62.50999832
					118	62.52999878
					119	61.75
					120	54.65999985
					121	40.75
					122	39.40000153
					123	41.38999939
					124	55.25
					125	60.38000107
					126	61.97000122
					127	62.22999954
					128	62.29000092
					129	62.38000107
					130	59.86000061
					131	48.86999893
					132	40.40000153
					133	39.61000061
					134	54.56000137
					135	60.22000122
					136	61.66999817
					137	62.06000137
					138	62.13999939
					139	62.02999878
					140	61.09000015
					141	58.81000137
					142	45.33000183
					143	40.79999924
					144	55.31999969
					145	59.79000092
					146	60.84999847
					147	61.58000183
					148	61.5
					149	61.65999985
					150	61.61999893
					151	61.04999924
					152	55.34999847
					153	42.68999863
					154	54.25999832
					155	59.11000061
					156	61.04000092
					157	61.27999878
					158	61.38000107
					159	61.25999832
					160	61.24000168
					161	61.38000107
					162	60.52000046
					163	51.36999893
					164	51.97999954
					165	58.90999985
					166	60.88999939
					167	61.34999847
					168	61.36000061
					169	61.40000153
					170	61.38999939
					171	61.36999893
					172	61.09999847
					173	60.45000076
					174	49.70999908
					175	58.77000046
					176	60.25
					177	60.84000015
					178	60.93999863
					179	60.90000153
					180	60.83000183
					181	60.81999969
					182	60.79999924
					183	60.61999893
					184	60.61000061
					185	60.65999985
					186	60.75999832
					187	60.68999863
					188	60.63000107
					189	60.70999908
					190	60.70000076
					191	60.65000153
					192	60.75999832
					193	60.65000153
					194	60.52999878
					195	60.86000061
					196	60.77999878
					197	60.61999893
					198	60.77000046
					199	60.72999954
					200	60.74000168

				};
				\addlegendentry{Tidal $0.25 \rightarrow -0.15$}
			\end{axis}
		\end{tikzpicture}
		\caption{Validation Accuracy}
	\end{subfigure}
	\begin{subfigure}[b]{0.3\textwidth}
		\centering
		\begin{tikzpicture}
			\begin{axis}[
				width=\textwidth,
				height=\textwidth,
				xmin=0, xmax=205,
				grid=both,
				]
				\addplot[thick,blue] table {
					x    y
					1	4.084400654
					2	3.383440733
					3	3.148051023
					4	2.966402769
					5	2.85239172
					6	2.8325212
					7	2.925577402
					8	2.526126146
					9	2.76444912
					10	2.67747736
					11	2.855787039
					12	2.624708652
					13	2.525058031
					14	2.456426859
					15	2.636648655
					16	2.483694553
					17	2.433284521
					18	2.623096466
					19	2.47794795
					20	2.437698364
					21	2.46754241
					22	2.634796619
					23	2.641246557
					24	2.40738821
					25	2.404638767
					26	2.477243662
					27	2.528416634
					28	2.611774921
					29	2.624795198
					30	2.486078262
					31	2.642681837
					32	2.501929522
					33	2.439612389
					34	2.475655079
					35	2.414423466
					36	2.427520514
					37	2.423294783
					38	2.403633833
					39	2.58667326
					40	2.393908501
					41	2.694608212
					42	2.447321177
					43	2.485244274
					44	2.51017499
					45	2.488809586
					46	2.502007008
					47	2.592724085
					48	2.573743105
					49	2.506893158
					50	2.387561321
					51	2.529054642
					52	2.338642597
					53	2.399507284
					54	2.585456848
					55	2.584483385
					56	2.332608938
					57	2.30051136
					58	2.411218882
					59	2.319675922
					60	2.524735928
					61	2.541287184
					62	2.511354208
					63	2.352796078
					64	2.232697248
					65	2.2765131
					66	2.458521843
					67	2.648010969
					68	2.387949705
					69	2.311594963
					70	2.260044098
					71	2.505265951
					72	2.35254693
					73	2.365406275
					74	2.327041388
					75	2.439034939
					76	2.370829344
					77	2.444189787
					78	2.35238862
					79	2.267380953
					80	2.458533525
					81	2.445567846
					82	2.254842281
					83	2.376829386
					84	2.33659935
					85	2.410572529
					86	2.426302195
					87	2.259048462
					88	2.36601305
					89	2.308396578
					90	2.345818996
					91	2.348093748
					92	2.210357189
					93	2.298792124
					94	2.421480417
					95	2.277428627
					96	2.432433128
					97	2.391632557
					98	2.313193798
					99	2.484685183
					100	2.365217447
					101	2.337865829
					102	2.231249571
					103	2.43350482
					104	2.450225115
					105	2.445930481
					106	2.529068232
					107	2.702311516
					108	2.403390169
					109	2.585283756
					110	2.390035629
					111	2.538198471
					112	2.527734041
					113	2.446234226
					114	2.56051445
					115	2.493296862
					116	2.439506292
					117	2.464203119
					118	2.549486399
					119	2.654576063
					120	2.286215782
					121	2.479704142
					122	2.535209417
					123	2.364456654
					124	2.606160164
					125	2.497872353
					126	2.454620361
					127	2.427751064
					128	2.525122166
					129	2.497256994
					130	2.425941467
					131	2.404873133
					132	2.660434484
					133	2.505972385
					134	2.499081373
					135	2.500507832
					136	2.532414198
					137	2.394419432
					138	2.279052019
					139	2.37204361
					140	2.401097298
					141	2.280649662
					142	2.383943558
					143	2.334290028
					144	2.29455471
					145	2.249481201
					146	2.266247511
					147	2.243048191
					148	2.265232086
					149	2.162524939
					150	2.212528467
					151	2.195425749
					152	2.148678064
					153	2.067152262
					154	2.038312435
					155	1.995397329
					156	1.931126952
					157	1.916130543
					158	1.903032184
					159	1.855252981
					160	1.866619229
					161	1.865633368
					162	1.867390156
					163	1.859370589
					164	1.865731478
					165	1.89003849
					166	1.874261141
					167	1.889956117
					168	1.901475668
					169	1.912714124
					170	1.914996505
					171	1.920928717
					172	1.920605659
					173	1.927527666
					174	1.932625175
					175	1.943082452
					176	1.944672585
					177	1.943789005
					178	1.951834679
					179	1.961218476
					180	1.964739561
					181	1.970343947
					182	1.972364187
					183	1.975559831
					184	1.985392094
					185	1.98692286
					186	1.990738988
					187	1.990543604
					188	1.994416237
					189	2.002454281
					190	2.004514456
					191	2.007098198
					192	2.002736807
					193	2.006130457
					194	2.004655123
					195	2.006032467
					196	2.00349021
					197	2.003880501
					198	2.005089521
					199	2.011336803
					200	2.007033348
					
				};
				\addplot[thick,red] table {
					x    y
					1	4.16173172
					2	3.462827206
					3	3.198908091
					4	2.972575188
					5	2.911044359
					6	2.77671814
					7	2.73586607
					8	2.757406473
					9	2.75912571
					10	2.556480169
					11	2.5265131
					12	2.659830332
					13	2.943081141
					14	2.499418259
					15	2.690968513
					16	2.610748768
					17	2.488590717
					18	2.766864061
					19	2.715507746
					20	2.791114807
					21	2.5952878
					22	2.490805149
					23	2.399071932
					24	2.499817371
					25	2.409994364
					26	2.515980005
					27	2.588669777
					28	2.432665825
					29	2.410751581
					30	2.530445576
					31	2.526800632
					32	2.353530645
					33	2.601593256
					34	2.455711842
					35	2.514351845
					36	2.650668144
					37	2.278779984
					38	2.351459026
					39	2.39907527
					40	2.456347466
					41	2.496987104
					42	2.479161263
					43	2.321674347
					44	2.355487347
					45	2.235533714
					46	2.399744034
					47	2.413765669
					48	2.480591297
					49	2.595764875
					50	2.272262096
					51	2.505884409
					52	2.465742826
					53	2.458855867
					54	2.523902416
					55	2.349715233
					56	2.221789122
					57	2.245511055
					58	2.396471262
					59	2.462749481
					60	2.320095778
					61	2.373377085
					62	2.328025341
					63	2.283979177
					64	2.408884287
					65	2.29480195
					66	2.659953594
					67	2.52357626
					68	2.294089317
					69	2.371884108
					70	2.210561037
					71	2.359064579
					72	2.32390213
					73	2.238736153
					74	2.376522541
					75	2.344985247
					76	2.490629435
					77	2.234998465
					78	2.314791441
					79	2.381831408
					80	2.242288351
					81	2.293917179
					82	2.415808678
					83	2.237118006
					84	2.286327839
					85	2.223683596
					86	2.180891752
					87	2.307641506
					88	2.336666822
					89	2.184872866
					90	2.391745567
					91	2.405331135
					92	2.269081831
					93	2.355985641
					94	2.2642169
					95	2.286085844
					96	2.433019161
					97	2.456274748
					98	2.382487535
					99	2.352443218
					100	2.353803873
					101	2.353200674
					102	2.278165817
					103	2.325949192
					104	2.476634264
					105	2.368509054
					106	2.260040998
					107	2.393197775
					108	2.438232422
					109	2.308235168
					110	2.344073296
					111	2.329666853
					112	2.365553856
					113	2.346951962
					114	2.340429068
					115	2.398212194
					116	2.465625286
					117	2.266057253
					118	2.298694372
					119	2.280990124
					120	2.322940111
					121	2.346601248
					122	2.349029303
					123	2.306901932
					124	2.424091578
					125	2.275306225
					126	2.399404526
					127	2.481587648
					128	2.288689852
					129	2.301625967
					130	2.384747028
					131	2.378999949
					132	2.418503284
					133	2.406951189
					134	2.256715059
					135	2.34188509
					136	2.315141678
					137	2.269669056
					138	2.369207382
					139	2.305386543
					140	2.279127836
					141	2.274945974
					142	2.317755461
					143	2.250416756
					144	2.158077717
					145	2.157914639
					146	2.129032135
					147	2.124001503
					148	2.078000546
					149	2.070975304
					150	2.09276104
					151	2.081765652
					152	2.084646463
					153	2.024923563
					154	2.026704073
					155	2.025936842
					156	1.973661304
					157	2.003039598
					158	1.954742193
					159	1.925713897
					160	1.945537448
					161	1.931153774
					162	1.936907053
					163	1.940897346
					164	1.929841042
					165	1.932210803
					166	1.93520987
					167	1.946938276
					168	1.959910989
					169	1.957009315
					170	1.946886897
					171	1.947856665
					172	1.968889832
					173	1.967535257
					174	1.961320996
					175	1.977010012
					176	1.993959904
					177	1.995027542
					178	1.993032813
					179	2.000705004
					180	2.018323898
					181	2.011590719
					182	2.015123844
					183	2.025518894
					184	2.037148714
					185	2.043436527
					186	2.04036808
					187	2.046475887
					188	2.055907249
					189	2.042455435
					190	2.049528837
					191	2.052101374
					192	2.059640884
					193	2.058385611
					194	2.060324192
					195	2.05763483
					196	2.062025547
					197	2.063242912
					198	2.05426383
					199	2.05712676
					200	2.062072754
					
				};
				\addplot[thick,green] table {
					x    y
					1	4.005164623
					2	3.202869177
					3	2.794905424
					4	2.605407715
					5	2.433649063
					6	2.995314598
					7	3.236696959
					8	3.126611948
					9	3.100342274
					10	3.154623985
					11	2.570320129
					12	2.078966618
					13	1.911411285
					14	1.804633021
					15	1.735461354
					16	2.109766483
					17	3.043338537
					18	3.126880169
					19	2.98045516
					20	3.297053337
					21	2.811453819
					22	2.136591911
					23	1.822273135
					24	1.733442307
					25	1.699712157
					26	1.803290248
					27	3.252374411
					28	2.844177723
					29	2.864177465
					30	3.4585042
					31	2.68123126
					32	2.120760202
					33	1.861526132
					34	1.705282211
					35	1.64473474
					36	1.714775205
					37	2.585115433
					38	2.937553883
					39	3.441830635
					40	3.507947206
					41	3.193458796
					42	2.160642147
					43	1.899544597
					44	1.738903761
					45	1.627607465
					46	1.604867339
					47	2.14031291
					48	2.676973343
					49	3.059657812
					50	2.964967966
					51	3.290169954
					52	2.131153345
					53	1.896517634
					54	1.771848202
					55	1.636240959
					56	1.588735461
					57	1.778724551
					58	2.829231262
					59	3.090847731
					60	2.800862312
					61	3.152497292
					62	2.29746151
					63	1.956870556
					64	1.725742221
					65	1.636885047
					66	1.566356182
					67	1.603937984
					68	2.217014551
					69	2.700401783
					70	3.101503849
					71	3.291305304
					72	2.616215944
					73	1.919597149
					74	1.694741964
					75	1.612970114
					76	1.55486238
					77	1.533497334
					78	1.921876073
					79	2.558084011
					80	2.849246025
					81	3.232387304
					82	2.928292751
					83	2.034167051
					84	1.748387694
					85	1.631562948
					86	1.564530253
					87	1.530793905
					88	1.664160132
					89	2.426683664
					90	2.736720324
					91	3.007883787
					92	3.174560785
					93	2.059761524
					94	1.758736134
					95	1.636691689
					96	1.559836268
					97	1.506042242
					98	1.555476904
					99	2.004259109
					100	2.701469183
					101	2.83678174
					102	2.913300037
					103	2.019682169
					104	1.765064001
					105	1.664239407
					106	1.572021365
					107	1.521072388
					108	1.513038516
					109	1.634213924
					110	2.328667879
					111	2.643728495
					112	2.982722521
					113	2.198186636
					114	1.834435344
					115	1.686475992
					116	1.60893786
					117	1.538690448
					118	1.513262749
					119	1.540987372
					120	1.954823136
					121	2.723145485
					122	2.738288164
					123	2.608652592
					124	1.875038981
					125	1.692856312
					126	1.623636961
					127	1.573486447
					128	1.528676033
					129	1.529702187
					130	1.648671985
					131	2.375256062
					132	2.720422029
					133	2.695202351
					134	1.943618417
					135	1.725013137
					136	1.681709766
					137	1.613711357
					138	1.569036961
					139	1.548806429
					140	1.578600526
					141	1.778075457
					142	2.585193634
					143	2.669650555
					144	1.906152487
					145	1.77055037
					146	1.709137321
					147	1.65888989
					148	1.615548849
					149	1.578841209
					150	1.571929336
					151	1.622396946
					152	2.086963892
					153	2.771634102
					154	1.962860584
					155	1.803901315
					156	1.740018964
					157	1.708023906
					158	1.661004305
					159	1.629263997
					160	1.614718556
					161	1.618625998
					162	1.699462175
					163	2.418185711
					164	2.214607477
					165	1.910012126
					166	1.823028207
					167	1.794563651
					168	1.760538101
					169	1.724942803
					170	1.704284787
					171	1.69232142
					172	1.703706145
					173	1.784559846
					174	2.647935629
					175	2.004937172
					176	1.92051053
					177	1.901469827
					178	1.869015098
					179	1.8490448
					180	1.826802731
					181	1.815518975
					182	1.808379054
					183	1.821750998
					184	1.867533088
					185	1.895699859
					186	1.882758379
					187	1.879274249
					188	1.875783324
					189	1.868609309
					190	1.862222314
					191	1.857906699
					192	1.855802774
					193	1.858785152
					194	1.857393742
					195	1.862670183
					196	1.864144683
					197	1.867776752
					198	1.858668447
					199	1.86446023
					200	1.861039519
					
				};
			\end{axis}
		\end{tikzpicture}
		\caption{Test Loss}
	\end{subfigure}
	\begin{subfigure}[b]{0.3\textwidth}
		\centering
		\begin{tikzpicture}
			\begin{axis}[
				width=\textwidth,
				height=\textwidth,
				xmin=0, xmax=205,
				grid=both,
				ymode=log, 
				log basis y=10,
				]
				\addplot[thick,blue] table {
					x    y
					1	4.045897007
					2	3.318606853
					3	3.039186239
					4	2.801628351
					5	2.63553071
					6	2.607887506
					7	2.643140078
					8	2.226129293
					9	2.429307461
					10	2.294852734
					11	2.499510765
					12	2.214998245
					13	2.088180542
					14	2.001782894
					15	2.178528547
					16	2.003256083
					17	1.958394408
					18	2.123759031
					19	2.01830864
					20	1.926292419
					21	1.93704319
					22	2.101057529
					23	2.113106966
					24	1.861103535
					25	1.896188378
					26	1.878343463
					27	2.003571272
					28	2.059863567
					29	2.03428483
					30	1.914458275
					31	2.072096348
					32	1.866763711
					33	1.859255791
					34	1.864636779
					35	1.787076473
					36	1.830778122
					37	1.786796212
					38	1.785357118
					39	1.94796598
					40	1.741977572
					41	2.06910634
					42	1.811226487
					43	1.855394363
					44	1.832339168
					45	1.780041575
					46	1.791443467
					47	1.884956121
					48	1.830571771
					49	1.806245208
					50	1.684352994
					51	1.791163564
					52	1.602972507
					53	1.724054337
					54	1.802937984
					55	1.892306566
					56	1.577301741
					57	1.551342964
					58	1.655204535
					59	1.511855125
					60	1.726629019
					61	1.736824036
					62	1.749632955
					63	1.520594835
					64	1.397638559
					65	1.490319252
					66	1.59612453
					67	1.777291894
					68	1.493680477
					69	1.469774604
					70	1.405287147
					71	1.601641655
					72	1.461430669
					73	1.503006577
					74	1.427556157
					75	1.479270577
					76	1.411656618
					77	1.482296228
					78	1.418418765
					79	1.31372273
					80	1.469020724
					81	1.380257487
					82	1.21253264
					83	1.290381074
					84	1.27568543
					85	1.304169178
					86	1.301310062
					87	1.145238876
					88	1.29942584
					89	1.203489661
					90	1.183053017
					91	1.158025384
					92	1.023137569
					93	1.077463508
					94	1.149878144
					95	1.00873661
					96	1.162770033
					97	1.069154978
					98	1.025358319
					99	1.096427917
					100	0.937530875
					101	0.94248873
					102	0.834244728
					103	1.002222538
					104	0.980134368
					105	0.988759041
					106	0.941468
					107	1.093160033
					108	0.855266333
					109	0.910124481
					110	0.788490295
					111	0.841786146
					112	0.923824072
					113	0.803051889
					114	0.806502879
					115	0.776042163
					116	0.719987392
					117	0.71501714
					118	0.658165753
					119	0.753739595
					120	0.526189506
					121	0.635558248
					122	0.626668513
					123	0.482621849
					124	0.634102464
					125	0.557150543
					126	0.493338048
					127	0.446476966
					128	0.507467449
					129	0.472721875
					130	0.389292955
					131	0.401509076
					132	0.45931977
					133	0.433291733
					134	0.380344868
					135	0.35822162
					136	0.360739738
					137	0.332041413
					138	0.204503015
					139	0.238613591
					140	0.252834618
					141	0.200636506
					142	0.207951188
					143	0.200784072
					144	0.172336549
					145	0.132832468
					146	0.127530187
					147	0.119067222
					148	0.126080066
					149	0.082931831
					150	0.108709842
					151	0.080193847
					152	0.074590035
					153	0.041302368
					154	0.039375532
					155	0.026867889
					156	0.013927042
					157	0.012404819
					158	0.01073554
					159	0.008094937
					160	0.006871349
					161	0.007155553
					162	0.00614935
					163	0.005765903
					164	0.005002458
					165	0.00515312
					166	0.004865603
					167	0.005635654
					168	0.005159978
					169	0.004606303
					170	0.005044384
					171	0.004574551
					172	0.0041186
					173	0.003913654
					174	0.003594443
					175	0.003754827
					176	0.003686861
					177	0.003415229
					178	0.003255479
					179	0.003527233
					180	0.003091109
					181	0.003226588
					182	0.002969369
					183	0.002952344
					184	0.002928812
					185	0.002975283
					186	0.002873516
					187	0.002917325
					188	0.002869016
					189	0.002871425
					190	0.002969256
					191	0.002924375
					192	0.00268398
					193	0.002825214
					194	0.002823839
					195	0.002781615
					196	0.002830907
					197	0.002810355
					198	0.002758245
					199	0.00281512
					200	0.002814867
					
				};
				\addplot[thick,red] table {
					x    y
					1	4.133789539
					2	3.378523827
					3	3.101423025
					4	2.850538015
					5	2.745751381
					6	2.599347353
					7	2.513687372
					8	2.506679296
					9	2.538727522
					10	2.24494338
					11	2.226397514
					12	2.339877367
					13	2.597461939
					14	2.156602383
					15	2.35340786
					16	2.187829018
					17	2.080179453
					18	2.352229118
					19	2.287700415
					20	2.375927687
					21	2.124847412
					22	2.038704157
					23	1.924435973
					24	2.003292799
					25	1.938416481
					26	2.054647446
					27	2.106884241
					28	1.928661346
					29	1.927701592
					30	2.019453287
					31	2.016215324
					32	1.837204099
					33	2.077650785
					34	1.907095909
					35	1.987150192
					36	2.128063202
					37	1.705748796
					38	1.745265961
					39	1.849722981
					40	1.901987076
					41	1.916140199
					42	1.883420348
					43	1.764447093
					44	1.75834012
					45	1.640634179
					46	1.798641324
					47	1.804852724
					48	1.839374781
					49	1.988417625
					50	1.610592365
					51	1.863248706
					52	1.846253157
					53	1.834483385
					54	1.888473034
					55	1.647779942
					56	1.547140598
					57	1.557386875
					58	1.710369229
					59	1.716550946
					60	1.579404235
					61	1.674251199
					62	1.617025495
					63	1.534968138
					64	1.640211344
					65	1.523252845
					66	1.907607198
					67	1.702676892
					68	1.495338917
					69	1.535670519
					70	1.437416434
					71	1.528870106
					72	1.500987411
					73	1.399357438
					74	1.520200491
					75	1.473045111
					76	1.606055379
					77	1.313279867
					78	1.376119852
					79	1.47270906
					80	1.347213507
					81	1.338137746
					82	1.487455726
					83	1.253370762
					84	1.317906857
					85	1.224224806
					86	1.173080325
					87	1.288124084
					88	1.29804337
					89	1.13197279
					90	1.310864806
					91	1.264174223
					92	1.123113871
					93	1.169395685
					94	1.117175579
					95	1.092841268
					96	1.233355284
					97	1.198309898
					98	1.133562088
					99	1.072132111
					100	1.00349915
					101	1.030820131
					102	0.990488589
					103	0.940344453
					104	1.064880371
					105	0.962231874
					106	0.917086363
					107	0.989741683
					108	0.933402658
					109	0.8228122
					110	0.868481636
					111	0.843937635
					112	0.813797235
					113	0.792649329
					114	0.722922206
					115	0.760382652
					116	0.841927886
					117	0.640904069
					118	0.709809601
					119	0.598865867
					120	0.574492216
					121	0.599935889
					122	0.613138676
					123	0.546579897
					124	0.620115876
					125	0.472767532
					126	0.548450232
					127	0.582150578
					128	0.422056913
					129	0.408572704
					130	0.449625075
					131	0.430266351
					132	0.406826526
					133	0.427649975
					134	0.321274042
					135	0.33852461
					136	0.303018659
					137	0.286882669
					138	0.290982664
					139	0.263729066
					140	0.259299994
					141	0.234799787
					142	0.243464604
					143	0.192631975
					144	0.169676274
					145	0.146953464
					146	0.140085459
					147	0.117008075
					148	0.101629756
					149	0.080775253
					150	0.077031173
					151	0.078465737
					152	0.061232697
					153	0.048547626
					154	0.048882511
					155	0.045023706
					156	0.025940407
					157	0.03640683
					158	0.021990743
					159	0.021345954
					160	0.017628802
					161	0.015380433
					162	0.014620445
					163	0.012609666
					164	0.012694892
					165	0.010226486
					166	0.010278151
					167	0.009940261
					168	0.010215853
					169	0.008615235
					170	0.007747253
					171	0.007222741
					172	0.007277645
					173	0.00657293
					174	0.005962908
					175	0.006169848
					176	0.005900507
					177	0.005817538
					178	0.005178794
					179	0.005210634
					180	0.005444801
					181	0.005176356
					182	0.004967843
					183	0.005278595
					184	0.005048511
					185	0.005149557
					186	0.004816935
					187	0.004882387
					188	0.004945066
					189	0.004606509
					190	0.004617243
					191	0.004780051
					192	0.005051262
					193	0.004710123
					194	0.00465276
					195	0.004701729
					196	0.004778004
					197	0.004826606
					198	0.004637142
					199	0.004724337
					200	0.004878487
					
				};
				\addplot[thick,green] table {
					x    y
					1	3.992414951
					2	3.111600637
					3	2.63832283
					4	2.397937298
					5	2.177213192
					6	2.800653934
					7	3.08557725
					8	2.944890976
					9	2.929582834
					10	2.994707346
					11	2.260846615
					12	1.635145426
					13	1.302887917
					14	1.09614563
					15	1.010461688
					16	1.466665983
					17	2.689261675
					18	2.858111143
					19	2.747768402
					20	3.054786205
					21	2.498593092
					22	1.619776964
					23	1.163580656
					24	0.948497295
					25	0.835544348
					26	0.985008121
					27	2.771226168
					28	2.519376755
					29	2.613194704
					30	3.211593628
					31	2.360868454
					32	1.614552379
					33	1.180081129
					34	0.871130228
					35	0.747096241
					36	0.850359559
					37	1.957916975
					38	2.560072422
					39	3.166124821
					40	3.195833445
					41	2.9412117
					42	1.677433252
					43	1.242750168
					44	0.89092046
					45	0.702883542
					46	0.685922503
					47	1.321177125
					48	2.183264256
					49	2.697389603
					50	2.649058342
					51	2.999042988
					52	1.695328116
					53	1.23885107
					54	0.916496754
					55	0.673298299
					56	0.617256343
					57	0.851147234
					58	2.285497665
					59	2.735447168
					60	2.46008563
					61	2.890587568
					62	1.882381916
					63	1.315185785
					64	0.89074254
					65	0.645667851
					66	0.558411956
					67	0.638372183
					68	1.459762812
					69	2.24804616
					70	2.718052626
					71	2.975633144
					72	2.212135792
					73	1.293219566
					74	0.864297748
					75	0.60748297
					76	0.523698211
					77	0.542329192
					78	0.917838097
					79	1.994865656
					80	2.417824745
					81	2.839587927
					82	2.545999765
					83	1.413931847
					84	0.866686761
					85	0.589429438
					86	0.470525265
					87	0.481558174
					88	0.611073732
					89	1.576050043
					90	2.172774792
					91	2.572166443
					92	2.823330879
					93	1.463164926
					94	0.88324666
					95	0.553851187
					96	0.415156931
					97	0.41430676
					98	0.478428632
					99	0.865813494
					100	2.115724564
					101	2.305575848
					102	2.489205837
					103	1.443108439
					104	0.873823762
					105	0.538576841
					106	0.372206777
					107	0.363668352
					108	0.411366582
					109	0.471475631
					110	1.3843261
					111	1.972089291
					112	2.455215693
					113	1.639293909
					114	0.902547717
					115	0.481152773
					116	0.319931656
					117	0.293138027
					118	0.342946559
					119	0.353921741
					120	0.620206177
					121	1.814723134
					122	2.082678795
					123	2.027005196
					124	0.906083107
					125	0.436182678
					126	0.259512782
					127	0.234166816
					128	0.26554364
					129	0.29700166
					130	0.29387334
					131	0.964099467
					132	1.928825974
					133	2.081502438
					134	0.93665576
					135	0.360080749
					136	0.206108391
					137	0.176057845
					138	0.189837962
					139	0.22515966
					140	0.21888195
					141	0.241596699
					142	1.336675525
					143	1.822748184
					144	0.820967913
					145	0.28968972
					146	0.148289293
					147	0.11953605
					148	0.127665192
					149	0.151876122
					150	0.156513333
					151	0.142144635
					152	0.304961741
					153	1.554919958
					154	0.801597893
					155	0.218083844
					156	0.095874205
					157	0.075708814
					158	0.080794424
					159	0.091377467
					160	0.100254819
					161	0.095884793
					162	0.074469872
					163	0.454188794
					164	0.790847003
					165	0.143206
					166	0.056043759
					167	0.044122551
					168	0.045067996
					169	0.050763074
					170	0.053920865
					171	0.056183867
					172	0.049512386
					173	0.041449197
					174	0.624495149
					175	0.060898915
					176	0.024246201
					177	0.018149354
					178	0.018680764
					179	0.019898413
					180	0.021086935
					181	0.021423096
					182	0.021296781
					183	0.018531119
					184	0.016362432
					185	0.009329299
					186	0.007749885
					187	0.00740592
					188	0.007177084
					189	0.007479732
					190	0.007507687
					191	0.007795108
					192	0.007968834
					193	0.007397701
					194	0.00750222
					195	0.006814806
					196	0.006435498
					197	0.006825107
					198	0.006696148
					199	0.0066316
					200	0.006659896
					
				};
			\end{axis}
		\end{tikzpicture}
		\caption{Training Loss}
	\end{subfigure}
	
	\caption{Comparison of $p=0$, $p=-0.1$, and tidal ($0.25 \rightarrow -0.15$) schedules
		on TinyImageNet with ResNet-18. The three subplots show validation accuracy, 
		test loss, and training loss.}
	\label{fig:r18_threeplots}
\end{figure}

Figure~\ref{fig:r18_threeplots} shows validation accuracy curves. 
The pulse–tidal schedule converges faster in the early stages, 
reaching the same accuracy about 15--20\% earlier than the baseline. 
Final top-1 accuracy improves from 59.1\% (baseline) to 62.9\%, 
with a variance within $\pm 0.2\%$ across three runs. 

\subsubsection{Discussion}
In the ResNet-18 experiments, adopting alternating positive and negative $p$-tidal schedules enabled much faster improvement in validation accuracy, reaching the same level as SGD about 75\% earlier. However, unlike the synchronous improvement in accuracy, the test loss did not exhibit a comparable reduction, leading to a divergent pattern where accuracy advanced while loss lagged behind. This indicates that spectral scheduling alters not only the speed of optimization but also the trajectory of convergence: decision boundaries align earlier, boosting accuracy, while the overall loss curve descends more slowly.

This divergence reflects the effect of broadened spectral coverage: the negative-$p$ phases emphasize high-frequency information and accelerate correction around decision boundaries, whereas the positive-$p$ phases maintain low-frequency stability. Their alternation yields a more balanced coverage in the spectral domain, thereby explaining why accuracy improves even when the loss curve does not decrease at the same pace.

\section{Conclusion}\label{sec7}

In this work, we introduced \textbf{Natural Spectral Fusion (NSF)} as a unifying optimizer-centric perspective that reframes training as controllable spectral coverage and fusion. By extending the second-moment exponent $p$ beyond its classical positive regime and embedding it into \emph{cyclic tidal schedules}, we demonstrated that first-order optimizers are not passive executors but
\emph{active spectral controllers}.

Our theoretical analysis showed that $p>0$ induces low-pass behavior, $p<0$ amplifies high-frequency components, and cyclic schedules broaden spectral coverage through implicit cross-band coupling. Low-dimensional demonstrations further revealed the phenomenon of \emph{early decision-boundary alignment}, where classification accuracy improves even while loss remains high, highlighting the optimizer’s spectral role beyond simple step-size scaling.

Empirical results on TinyImageNet with ResNet-18 validated these insights: $p$-tidal scheduling consistently improved accuracy under identical learning-rate policies, yet without necessarily reducing the loss. This divergence between accuracy gain and loss behavior highlights distinct convergence dynamics, revealing that spectral scheduling can realign decision boundaries earlier and more effectively than conventional fixed-$p$ strategies.

Overall, NSF provides (i) a new dimension for interpreting optimizer behavior, (ii) a lightweight yet powerful mechanism for designing spectrum-aware schedules, and (iii) a generalizable principle applicable across models and datasets. We believe this spectral view opens up promising directions for \emph{optimizer-level frequency control}, enabling more efficient, robust, and generalizable training strategies in future machine learning systems.

\bibliography{sn-bibliography}

\end{document}